\documentclass[12pt,draftcls,onecolumn]{IEEEtran}
\usepackage{algorithm}
\usepackage{algorithmic}
\usepackage{subfigure}
\usepackage{graphicx,epsfig}
\usepackage{amsmath}
\usepackage{multirow}
\usepackage{enumerate}
\usepackage{bm}
\usepackage{multirow}
\usepackage{color}

\ifCLASSINFOpdf
\else
\fi

\begin{document}
%
\title{Joint Architecture and Knowledge Distillation in CNN for Chinese Text Recognition}

%

\author{Zi-Rui~Wang,
        and~Jun~Du
\thanks{Zi-Rui Wang, and Jun Du are with the National Engineering Laboratory for Speech and Language Information Processing, University of Science
and Technology of China, Hefei, Anhui, P. R. China (e-mail: cs211@mail.ustc.edu.cn; jundu@ustc.edu.cn).}}
\maketitle

\begin{abstract}
The distillation technique helps transform cumbersome neural networks into compact networks so that models can be deployed on alternative hardware devices. The main advantage of distillation-based approaches include a simple training process, supported by most off-the-shelf deep learning software and no special hardware requirements. In this paper, we propose a guideline for distilling the architecture and knowledge of pretrained standard CNNs. The proposed algorithm is first verified on a large-scale task: offline handwritten Chinese text recognition (HCTR). Compared with the CNN in the state-of-the-art system, the reconstructed compact CNN can reduce the computational cost by $>$10$\times$ and the model size by $>$8$\times$ with negligible accuracy loss. Then, by conducting experiments on {two additional classification task datasets: \textit{Chinese Text in the Wild} (CTW) and MNIST}, we demonstrate that the proposed approach can also be successfully applied on mainstream backbone networks.
\end{abstract}


\IEEEpeerreviewmaketitle

\section{Introduction}
%
%
%
%
\IEEEPARstart{C}{onvolutional} neural networks (CNNs) play an important role in the new wave of artificial intelligence. Since the first-generation CNNs were proposed by LeCun \cite{lecun1989backpropagation,lecun1998gradient} for handwritten character recognition, numerous CNNs have been emerging in different applications, such as (Alex, VGG, GoogLe, Res, and Dense)-Nets in natural image recognition \cite{krizhevsky2012imagenet,simonyan2014very,szegedy2015going,he2016deep,huang2017densely}, DCNN in offline handwritten Chinese text recognition (HCTR) \cite{wang2018comprehensive,wang2019writer}, HCCR-CNN in handwritten Chinese character recognition (HCCR) \cite{xiao2017building,zhang2017online}, FaceNet in face recognition \cite{schroff2015facenet} and FCN in speech emotion recognition \cite{zhang2018attention}. Such CNNs share the same basic components, i.e., convolutional layer, pooling layer and fully-connected layer. Although these networks have dramatically improved performance and CNN-based approaches have become mainstream in a wide range of pattern recognition tasks, the trend of going deeper and wider for CNNs makes them difficult to be deployed on resource-limited devices e.g., mobile phones and embedded chips. Moreover, there is an evident fact that the current state-of-the-art CNNs still mainly depend on massive handcrafted trail-and-error experiments. Both the architecture and the internal knowledge of a CNN should be valuable information for acceleration and compression algorithms. Accordingly, in this paper, we focus on both architecture and knowledge distillation in pretrained standard CNNs. 

The concept of knowledge distillation, which can be traced back to Caruana's research in 2006 \cite{bucilua2006model}, is to transfer the knowledge from cumbersome models into smaller model. Different from knowledge distillation, the research on architecture distillation in CNN focuses on inventing new efficient convolutions or units to directly replace standard convolutions of baseline CNN. A representative work was conducted in \cite{xu2019lightweightnet} where the authors reconstructed a lightweight CNN by using multiple efficient compact blocks according to the different locations of the baseline CNN. The realization of distillation can be figuratively described as teacher-student learning in which a network with massive parameters and high performance acts as a teacher and the compressed network is a student. Both the architecture and the internal knowledge of the teacher network should be learned by the student network. The single consideration usually leads to a contradiction between optimal performance and satisfactory compression. Unlike previous distillation algorithms, in this paper, we propose a guideline for distilling the architecture and knowledge of a pretrained CNN. Specifically, instead of using multiple acceleration blocks \cite{xu2019lightweightnet}, we develop a uniform block named the parsimonious convolution (ParConv) block that only consists of depthwise separable convolution (DSConv) \cite{chollet2017xception} and pointwise convolution in a heterogeneous combination. In knowledge distillation, a new solving procedure loss (SPL) is added to further improve the performance of the student network. The solving procedure is represented by the differences in attention maps between two layers.

The effectiveness of the proposed algorithm is mainly demonstrated on offline handwritten Chinese text recognition (HCTR). The HCTR can be widely used in many applications, such as mail address recognition \cite{fu2006novel}, bank check \cite{yu2001segmentation} and document recognition \cite{fujisawa2008forty}. Although the HCTR has made great progress owing to deep learning \cite{lecun2015deep}, it remains a challenging problem for the following reasons: 1) the text line must be considered as a whole rather than isolated characters, 2) more than 7,000 classes in common Chinese vocabulary and large-scale training samples, and 3) the unconstrained writing condition. The experiments are conducted on the ICDAR 2013 competition task of the CASIA-HWDB databse \cite{liu2011casia,yin2013icdar}, which is one of the most popular benchmark databases. To the best of our knowledge, no acceleration and compression approaches in CNNs have been validated on the offline HCTR. Furthermore, in order to display the generalization ability of the proposed algorithm, we use the proposed method to reduce the resource consumption of the mainstream backbone networks on CTW {\cite{yuan2018chinese}} and MNIST \cite{lecun1998gradient}.

The main contributions of this study can be summarized as follows:
\begin{itemize}
\item We propose a guideline for distilling the architecture and knowledge of pretrained standard CNNs for fast deployability on alternative hardware devices.
\item In architecture distillation, we invent a parsimonious convolution block (ParConv) to directly replace vanilla convolution without other adjustments. Compared with LightweightNet \cite{xu2019lightweightnet} and DSConv \cite{chollet2017xception}, the proposed ParConv demonstrates its superiority in recognition performance, computational cost and storage overhead.
\item In knowledge distillation, a new solving process loss (SPL) is added to further improve the performance of compressed CNN.
\item The effectiveness of the proposed approach is first verified on offline HCTR. No study has investigated whether previous acceleration and compression algorithms are still feasible in this field.
\item Compared with the baseline CNN in HCTR, our proposed joint architecture and knowledge distillation can reduce the computational cost by $>$10$\times$ and model size by $>$8$\times$ with negligible accuracy loss. Applying the algorithm to the mainstream backbone networks Res50 and Res18 \cite{he2016deep} on CTW and MNIST, respectively, both of the reconstructed compact networks can obtain an obvious reduction in resource consumption. Especially, the corresponding compact network of Res18 can obtain a $>$9$\times$ compression rate for both model size and computational cost with almost no decrease in accuracy.
\end{itemize}

The remainder of this paper is organized as follows. Section~\ref{sec:rw} reviews related work. Section~\ref{sec:details} elaborates on the details of the proposed approach. Section~\ref{sec:exp} reports the experimental results and analyses. Finally, Section~\ref{sec:con} concludes.

\section{Related Work}
\label{sec:rw}

\subsection{Acceleration and Compression} 
Almost all acceleration and compression algorithms can be divided into five groups: low-rank decomposition \cite{denton2014exploiting,jaderberg2014speeding,zhang2016accelerating,kim2015compression,ding2017compact}, parameter pruning \cite{han2015learning,guo2016dynamic,xiao2017building,liu2017learning,luo2017thinet,gordon2018morphnet}, parameter quantization \cite{han2015deep,vanhoucke2011improving,courbariaux2016binarized,li2016ternary,ding2017lightnn,ding2019flightnns} , compact network design \cite{iandola2016squeezenet,zhang2018shufflenet,ma2018shufflenet,howard2017mobilenets,sandler2018mobilenetv2} and distillation \cite{hinton2015distilling,romero2014fitnets,zagoruyko2016paying,yim2017gift,liu2019structured,he2019knowledge,xu2019lightweightnet,singh2019hetconv,chen2019drop}

As one of the first attempts for low-rank decompositions of filters, Denton et al. \cite{denton2014exploiting} proposed several decomposition designs along different dimensions. In \cite{jaderberg2014speeding}, the $k \times k$ filters were decomposed into $k \times 1$ and $1 \times k$ filters. A representative work comes from \cite{zhang2016accelerating}, where nonlinear units were considered in the decomposition algorithm based on the assumption that the filter response lies in a low-rank subspace. In \cite{kim2015compression,ding2017compact}, Tucker decomposition is used to achieve compression. Such algorithms need to be conducted layer by layer. Once a layer has been decomposed, the whole network is retrained by the backpropagation (BP) algorithm. For large-scale tasks, repeated decomposition and training are usually time consuming.

Parameter pruning is based on a reasonable idea that the low weights in a neural network are not important so that they can be safely removed. In \cite{han2015learning,guo2016dynamic}, the weights were kept or removed by comparison with a fixed threshold. Xiao et al. \cite{xiao2017building} proposed the adaptive drop-weight (ADW) to dynamically increase the threshold. Liu et al. \cite{liu2017learning} proposed channel sparsity regularization to prune channels with small scaling factors. At almost the same time, Luo et al. \cite{luo2017thinet} pruned filters based on the reconstruction error of the corresponding next layer by using a greedy algorithm. Fine-grained pruning \cite{han2015learning,xiao2017building} requires a special software/hardware accelerator. Although channel-level pruning \cite{liu2017learning,luo2017thinet} can be directly applied to existing software platforms, such as low-rank decomposition based algorithms, the requirement of repeated pruning and fine-tuning is time consuming for large-scale tasks. Besides, from recent research \cite{liu2018rethinking}, the pruned architecture, rather than a set of inherited important weights, is more crucial to the efficiency in the final model. Compared with layerwise pruning, a global pruning strategy \cite{gordon2018morphnet} might be more valuable.

For parameter quantization, by using the hash algorithm, Han et al. \cite{han2015deep} divided network weights into different groups and used the weights in the same group to share a value. Vanhoucke et al. \cite{vanhoucke2011improving} used an 8-bit type instead of the common 32-bit floating type in the network. Courbariaux et al. \cite{courbariaux2016binarized} proposed a binarized neural network in which all weights and outputs are constrained to \{1, -1\}, while Li et al. \cite{li2016ternary} quantized weights into \{-1, 0 ,1\}. Such methods can save a significant number of resources. However, the approach in \cite{han2015deep} requires additional space to store the original positions for shared weights, and low-bit approximation usually degrades network performance. Different from constraining the weights to +1 or -1, Ding et al. \cite{ding2017lightnn,ding2019flightnns} replaced the multiply-accumulate operation with one shift or a constrained number of shifts and adds, which can make trade-offs between accuracy and computational consumption.

An efficient and effective network structure can save a significant amount of memory and computational cost and yield competitive performance. Many compact blocks have been invented to control the fast increase in network parameters, such as the Fire module in SqueezeNet \cite{iandola2016squeezenet} and the Inception module in GoogLeNet \cite{szegedy2017inception}. The basic unit in these networks still consists of canonical convolution. As one low-consumption (storage and computational cost) substitute, depthwise separable convolution (DSConv) was first introduced in \cite{sifre2014rigid,chollet2017xception} and has become a key building block in recent compact networks \cite{zhang2018shufflenet,ma2018shufflenet,howard2017mobilenets,sandler2018mobilenetv2}. In addition, Guo et al. \cite{guo2018network} proved that DSConv is the principal components of standard convolution and can approximate the standard convolution in closed form. Although the DSConv is far more efficient than standard convolution, consistent observations can be found in \cite{chollet2017xception,xu2019lightweightnet}, simple replacement by using DSConv is not effective. Chollet et al. \cite{chollet2017xception} scaled up depthwise separable filters so that the DSConv-based network Xception with the same number of parameters as the Inception V3 \cite{szegedy2016rethinking} can outperform Inception V3. 


The first distillation in a neural network was completed by Hinton et al. in \cite{hinton2015distilling}, i.e., knowledge distillation. In \cite{hinton2015distilling}, the soft labels from multiple neural networks were used to guide the training of a single network. Soon after, Romero et al. \cite{romero2014fitnets} improved the algorithm of knowledge distillation by using the outputs of hidden layers and the soft labels from a shallow network with more parameters as hints to instruct a thin deep network. Recently, Zagoruyko et al. \cite{zagoruyko2016paying} attempted to transfer the defined attention map in convolutional layers from one network to another, inspired by the human visual experience. In \cite{yim2017gift}, the flow of solution procedure (FSP) matrix is defined to measure the change in information between two different layers for a compressed network to imitate the middle products of the baseline network. More recently, Liu et al. \cite{liu2019structured} integrated pixelwise loss, pairwise loss and generative adversarial loss in knowledge distillation for semantic segmentation. At almost the same time, under the framework of knowledge distillation, He et al. \cite{he2019knowledge} extracted more compact middle features by using a pretrained autoencoder and proposed an affinity distillation module to capture the long-range dependency. Both of them utilize the relationship between pixels, which is important to semantic segmentation. In contrast to the diversified knowledge distillation approaches, there are not many architecture distillation \cite{xu2019lightweightnet} approaches. Recent influential works about architecture distillation can be found in \cite{singh2019hetconv,chen2019drop}. They share a similar concept that the feature maps in the standard convolution layer are redundant. Singh et al. \cite{singh2019hetconv} proposed heterogeneous convolution (HetConv) with different kernel sizes in each layer to handle the corresponding parts of input feature maps while the feature maps in \cite{chen2019drop} were factorized into high frequency with fine spatial resolution and low frequency with smaller spatial size.


\subsection{Offline HCTR}
\begin{figure*}
\centering
\includegraphics[width=5.5in,height=2.6in]{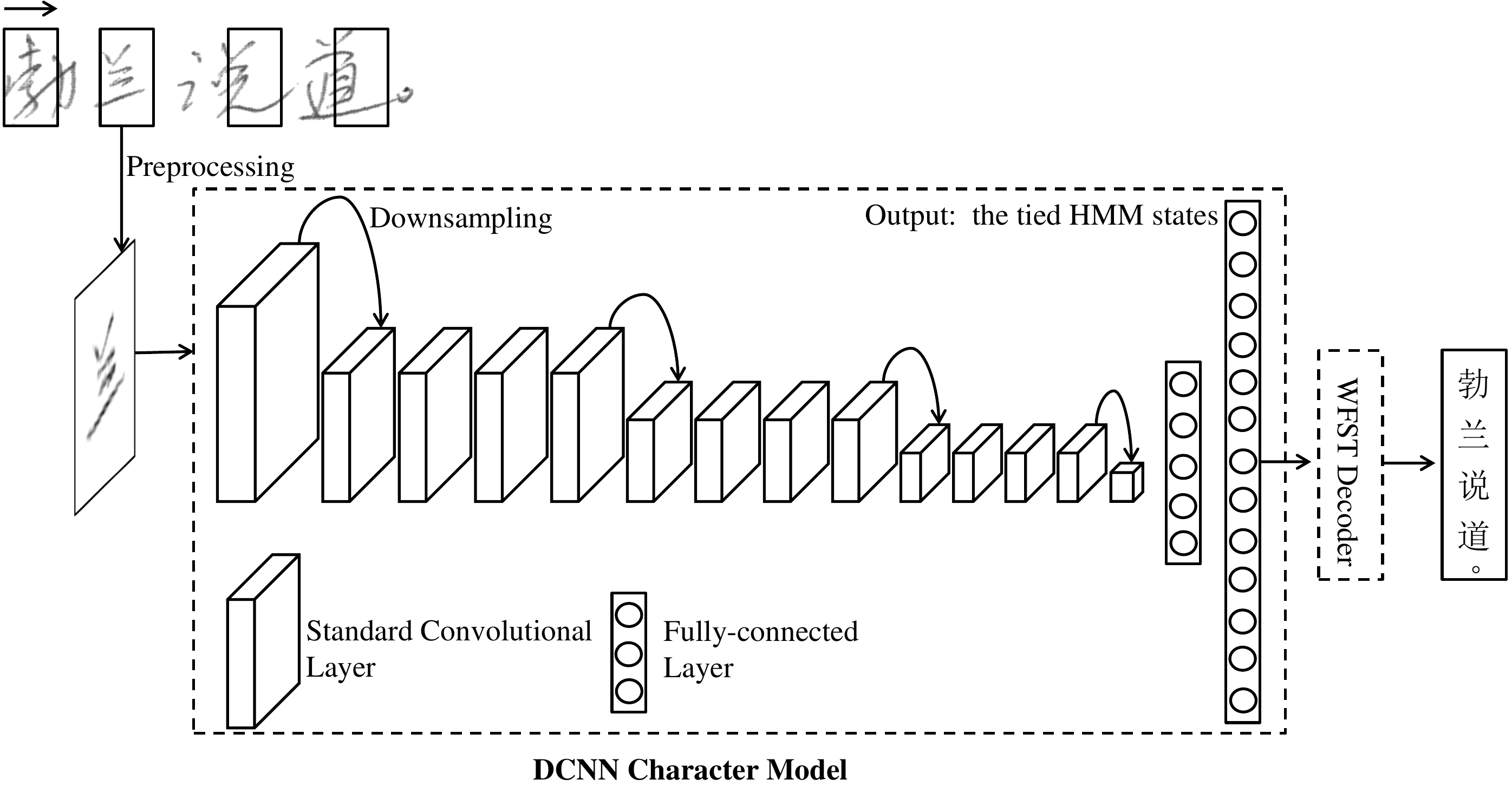}
\caption{Overview of our DCNN-based offline handwritten Chinese text recognition system.}
\label{DCNNSystem}
\end{figure*}

Offline HCTR can be formulated as a Bayesian decision problem:
\begin{eqnarray} \label{basic_principle}
\begin{aligned}
\displaystyle \hat{\mathbf{C}} &= \arg\max_{\mathbf{C}} p(\mathbf{C} | \mathbf{X})
\end{aligned}
\end{eqnarray}
where $\mathbf{X}$ is the feature sequence of a given text line image and $\mathbf{C}=\{C_1, C_2, ... , C_n\}$ is the underlying $n$-character sequence. The research efforts for addressing such sequence modeling tasks can be divided into three categories: oversegmentation \cite{wang2012handwritten,wang2016deep,wu2017improving}, connectionist temporal classification (CTC) \cite{messina2015segmentation,wu2017handwritten} and the hidden Markov model (HMM) \cite{su2009off,wang2018comprehensive,wang2019writer}. Almost all of these approaches benefit from the recent progress of deep learning \cite{lecun2015deep}. The outputs of neural networks in different modeling methods correspond to different concepts. For example, in oversegmentation and CTC-based approaches, the outputs of the neural network are related to segmentation identification or character classes. The outputs of the network used in HMM-based approaches are posterior probabilities of states. In our recent work \cite{wang2019writer}, each character is modeled by three tied states on average and a deep CNN (DCNN) with 22,080 (7,360$\times$3) output nodes is adopted as the character model and trained by hundreds of millions of frame-level images. As shown in Fig.~\ref{DCNNSystem}, frame-level images are extracted from original images by a left-to-right sliding window and fed into the DCNN. Then, the posterior probabilities of states are utilized in a WFST-based decoder \cite{mohri2002weighted} with/without language model (LM) for the final recognition results. In order to fit such massive training samples, the parameters of DCNN have been up to 124.5 MB and 16.02$\times $10$^8$  FLOPs are needed in each inference. More details and analyses of the DCNN are shown in Section~\ref{sec:exp}.


\section{Architecture and Knowledge Distillation}
\label{sec:details}
Given a baseline CNN(${{\bf{W}}_{{\rm{fc}}}},{{\bf{W}}_{{\rm{con}}}}$), let ${{\bf{W}}_{{\rm{fc}}}}$ represents the weight set of fully-connected layers and ${{\bf{W}}_{{\rm{con}}}}$ corresponds to the weight set of convolutional layers. For fully-connected layers in CNN, only the storage needs to be considered due to the relatively small computational cost. We use $\ell$ to denote the storage.  All parameters and calculations are based on a 32-bit floating point. Assuming the number of parameters is $M$, the storage  is computed as follows:

\begin{eqnarray}
\label{storage}
{l}{\rm{ = }} (\frac{{4M}}{{1,024 \times 1,024}}) {\kern 1pt} \rm{MB}
\end{eqnarray}
MB is the abbreviation for Mega Byte. The above equation is only the statistics for theoretical analysis. In experiments, the actual network storage is reported. The ratio $\gamma({{\bf{W}}_{{\rm{fc}}}},{{\bf{W}}_{{\rm{con}}}})  = \frac{{\ell ({{\bf{W}}_{{\rm{fc}}}})}}{{\ell ({{\bf{W}}_{{\rm{fc}}}}) + \ell ({{\bf{W}}_{\rm{con}}})}}$ is used to measure whether a strategy $\pi$ is conducted on the weights of fully-connected layers or not.

\begin{figure*}
\centering
\includegraphics[width=5.8in,height=3.8in]{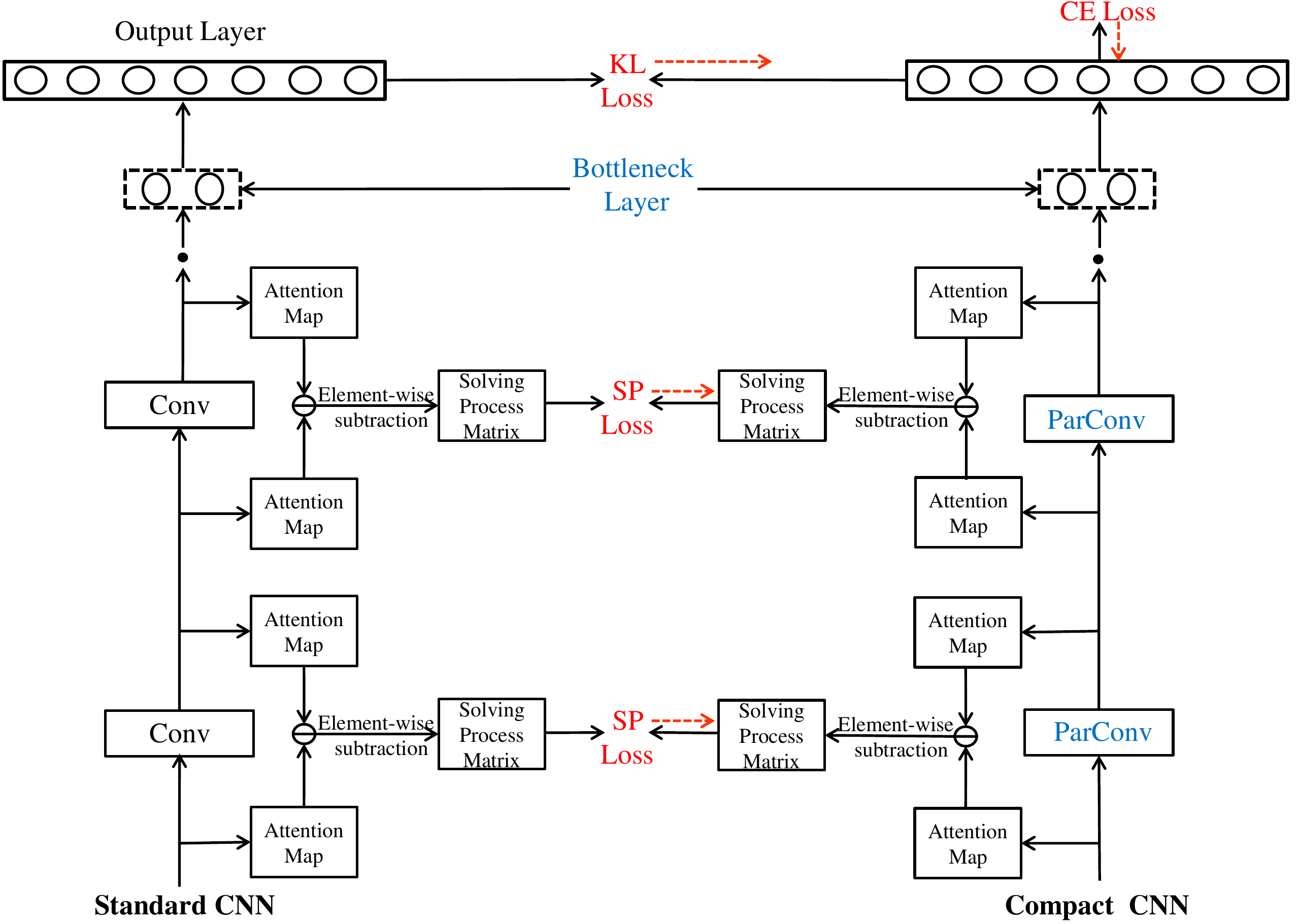}
\caption{The simplified framework of joint architecture and knowledge distillation.}
\label{Dis}
\end{figure*}

As summarized in Algorithm~\ref{alg:algorithm_dis_train}, the guideline involves architecture distillation and knowledge distillation. We first analyze the computational cost and storage overhead of the baseline CNN(${{\bf{W}}_{{\rm{fc}}}},{{\bf{W}}_{{\rm{con}}}}$) and compute the corresponding $\gamma({{\bf{W}}_{{\rm{fc}}}},{{\bf{W}}_{{\rm{con}}}})$. Then, in architecture distillation, if the weights of fully-connected layers occupy non-ignorable consumption of a certain computing resource (i.e., storage), we find a strategy $\pi({{\bf{W}}_{{\rm{fc}}}})$ to construct a new CNN(${\pi({\bf{W}_{{\rm{fc}}}})},{{\bf{W}}_{{\rm{con}}}}$) and ensure $\gamma(\pi({{\bf{W}}_{{\rm{fc}}}}),{{\bf{W}}_{{\rm{con}}}}) \leq T$ with the neglected performance loss (even better). In most cases, it is easy to find such a strategy for fully-connected layers, e.g., global pooling \cite{iandola2016squeezenet}, low-rank decomposition \cite{sainath2013low},and low-dimensional features \cite{xu2019lightweightnet}. Because we mainly focus on the compression of convolutional layers in this study, a naive solution $\pi$ that depends on the number of active output targets \cite{sainath2013low} to find an appropriate bottleneck feature before the output layer is adopted. For convolutional layers, the proposed ParConv blocks are used as a direct replacement to build a compact CNN (CCNN). Finally, in order to maintain the performance of the CCNN, knowledge distillation with three kinds of losses, namely, the Kullback-Leibler (KL) divergence loss, the cross entropy (CE) loss and solving process (SP) loss, is adopted to transfer knowledge from the standard CNN into the ParConv-based CCNN. Fig.~\ref{Dis} illustrates the proposed algorithm. More details of the respective parts are described in the following subsections.

\begin{algorithm}[htb]
\caption{The guideline of joint architecture and knowledge distillation.}
\label{alg:algorithm_dis_train}
\begin{algorithmic}[1]
\REQUIRE ~~\\
Baseline CNN(${{\bf{W}}_{{\rm{fc}}}},{{\bf{W}}_{{\rm{con}}}}$).\\
Threshold $T$
\STATE Analyze the computational cost and storage overhead in baseline CNN(${{\bf{W}}_{{\rm{fc}}}},{{\bf{W}}_{{\rm{con}}}}$) and compute $\gamma({{\bf{W}}_{{\rm{fc}}}},{{\bf{W}}_{{\rm{con}}}})$.
\IF{$\gamma({{\bf{W}}_{{\rm{fc}}}},{{\bf{W}}_{{\rm{con}}}}) > T$}
\STATE Find a strategy $\pi({{\bf{W}}_{{\rm{fc}}}})$ to construct a new CNN(${\pi({\bf{W}_{{\rm{fc}}}})},{{\bf{W}}_{{\rm{con}}}}$) with $\gamma(\pi({{\bf{W}}_{{\rm{fc}}}}),{{\bf{W}}_{{\rm{con}}}}) \leq T$ and neglected performance loss (even better).
\ENDIF
\STATE Build a CCNN by using ParConv blocks to replace the convolutional layers in the CNN.
\STATE Distill the knowledge of the CNN into the CCNN.
\RETURN The CCNN
\end{algorithmic}
\end{algorithm}

\begin{figure}
\centering
\includegraphics[width=3in,height=3in]{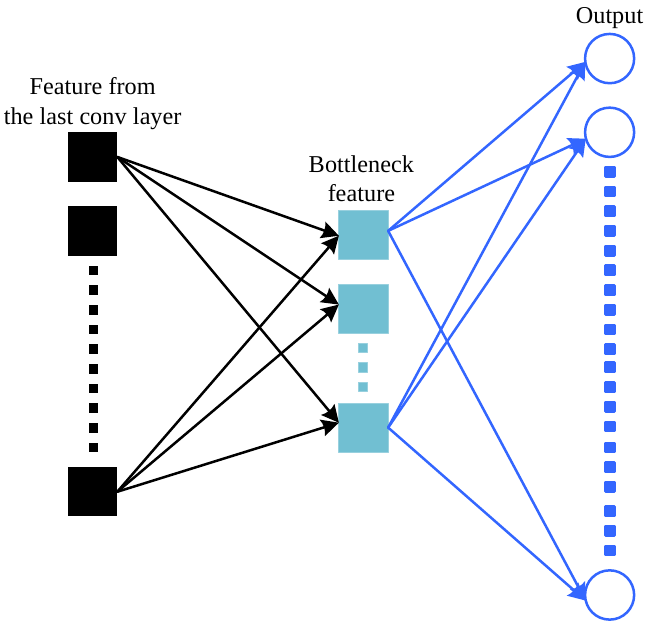}
\caption{Using a bottleneck layer to reduce the parameters of fully-connected layers.}`
\label{BNF}
\end{figure}

\subsection{Bottleneck Feature}
As shown in Fig.~\ref{BNF}, if there are $M$-dimensional features from the last conv layer, $B$-dimensional bottleneck features and $O$ output nodes, the total computational costs (FLOPs) of fully-connected layers (FCs) is computed as :
\begin{eqnarray}
\label{dsconv}
F{L_{{\rm{FCs}}}} = M \times B + B \times O
\end{eqnarray}
From the above equation, we can observe that the FLOPs in fully-connected layers can be controlled by adjusting the dimension of the bottleneck feature. The FLOPs of fully-connected layers becomes smaller with the reduction of the dimension $B$.


\subsection{Parsimonious Convolution}
In a standard convolutional layer, assuming the input is a square feature map, it can be represented by a three-dimensional tensor of size $D \times D \times C_{\rm{in}}$. Here, $D$ is the spatial width and height, while $C_{\rm{in}}$ is the number of input channels. Usually, (e.g., 3 $\times$ 3 kernel size and 1 padding), the corresponding output tensor with the channels $C_{\rm{out}}$ obtained by applying the $C_{\rm{out}}$ filters of size $K \times K \times C_{\rm{in}}$ has the same spatial size $D \times D$, namely, the output size is $D \times D \times C_{\rm{out}}$. Therefore, the FLOPs at this layer is:

\begin{eqnarray}
\label{conv}
F{L_{{\rm{Conv}}}} = D^2 \times {C_{{\rm{in}}}} \times {C_{{\rm{out}}}} \times K^2
\end{eqnarray}

The depthwise separable convolution (DSConv) is made up of two components: channelwise convolution and pointwise convolution. The fundamental hypothesis behind DSConv is that cross-channel correlations and spatial correlations can be decoupled. Channelwise convolution is used to capture spatial correlations and pointwise convolution is a $1 \times 1$ standard convolution that combines information from different channels. In channelwise convolution, each output channel is only associated with one input channel so that the convolutional filters are represented by a 3-D tensor $K \times K \times C_{\rm{in}}$. The FLOPs of DSConv is computed as follows:

\begin{eqnarray}
\label{dsconv}
F{L_{{\rm{DSConv}}}} = {\kern 1pt} D^2 \times ({C_{{\rm{in}}}} \times K^2 + {C_{{\rm{in}}}} \times {C_{{\rm{out}}}})
\end{eqnarray}

Compared with standard convolution, DSConv is extremely efficient in building units for many compact networks \cite{zhang2018shufflenet,ma2018shufflenet,howard2017mobilenets,sandler2018mobilenetv2}. However, directly replacing standard convolution with DSConv leads to an increase in network depth, which makes the optimization of the reconstructed network more difficult. This problem might be alleviated by using residual connections. Besides, simple replacement by using DSConv in a standard CNN causes the network to suffer performance degradation, which may be the reason that the authors in \cite{chollet2017xception} had to scale up the number of filters in DSConv.




In the proposed parsimonious convolution (ParConv), the input channels are split into two branches, one with $\alpha {C_{{\rm{in}}}}$ ($0 \leq \alpha \leq 1$) channels for DSConv and the other with $(1 - \alpha ){C_{{\rm{in}}}}$ channels for pointwise convolution. And then, the output features from different branches combine together by channel-wise addition. Specifically, before DSConv, a pointwise convolution with a channel multiplier $\omega$ is added to deeply integrate the information among channels, which is important for DSConv to extract features. In order to promote the flow of information between branches, a channel shuffle operator \cite{zhang2018shufflenet} is conducted before the input feature maps are split into two branches. The channel shuffle operator first reshapes the input channel dimension into ($2$, $\frac{C_{{\rm{in}}}}{2}$), transposing and then flattening it back. For simplicity, $\alpha$ is set to 0.5 in all ParConvs. The FLOPs of ParConv is:

\begin{eqnarray}
\begin{split}
F{L_{{\rm{ParConv}}}} =&  D^2 \times  \frac{1}{2}{C_{{\rm{in}}}} \times \frac{\omega }{2}{C_{{\rm{in}}}}+ D^2\times \frac{\omega }{2}{C_{{\rm{in}}}} \times K^2   \\
& + D^2\times\frac{\omega }{2}{C_{{\rm{in}}}} \times {C_{\rm{out}}}  + D^2 \times \frac{1}{2}{C_{{\rm{in}}}} \times {C_{\rm{out}}}
\end{split}
\end{eqnarray}

\begin{figure*}
\centering
\includegraphics[width=4.8in,height=2.4in]{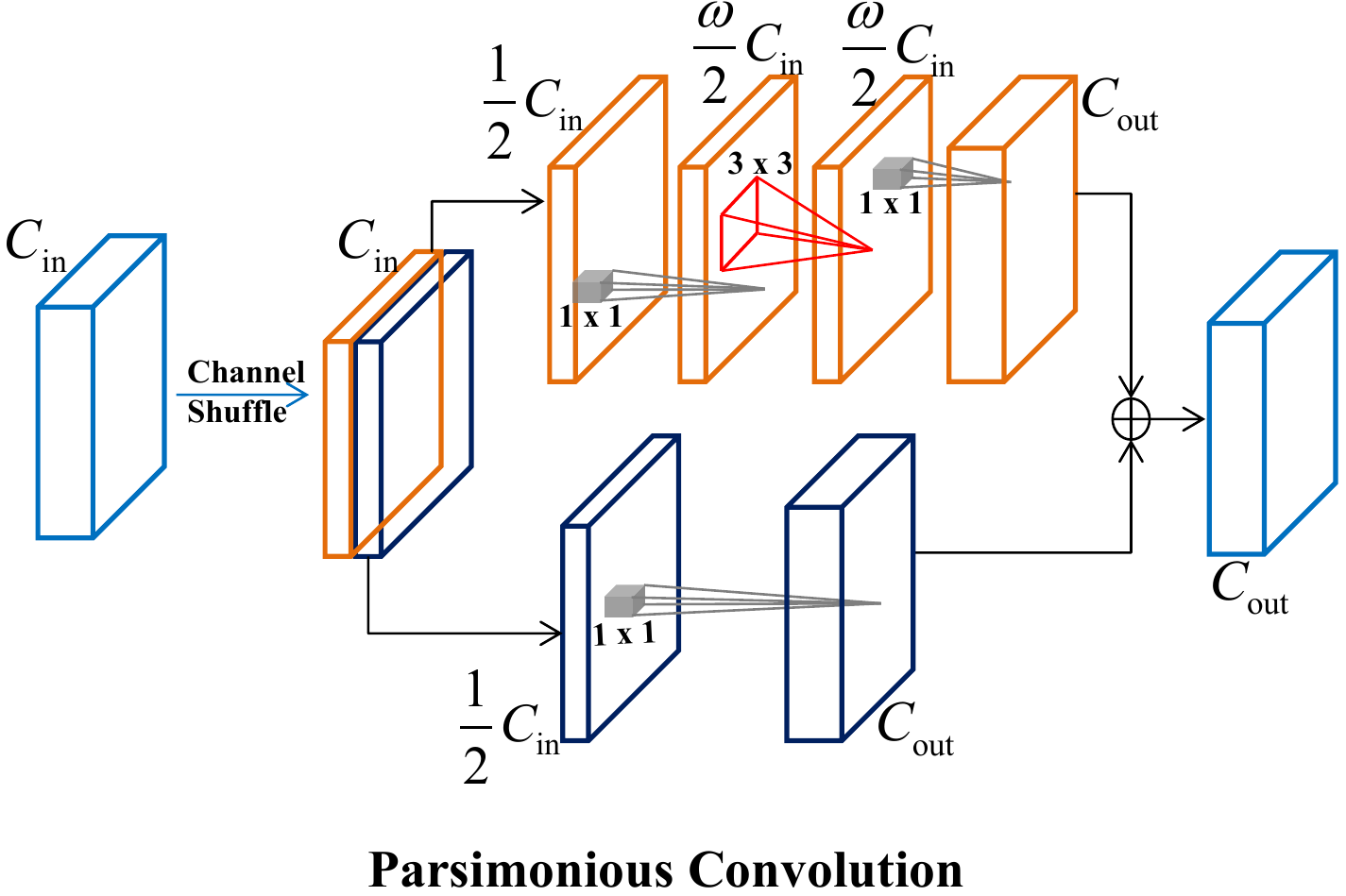}
\caption{The structure of the ParConv. The solid cube indicates that the kernel connects with all input channels, while the rectangle with a triangle indicates the kernel only works on the corresponding single input channel.}
\label{Conv}
\end{figure*}

Fig.~\ref{Conv} shows the structure of the ParConv and Table~\ref{RadioFLOPs} lists the FLOPs ratios of different compact convolutions to standard convolution. From Table~\ref{RadioFLOPs}, it can be observed clearly that the DSConv can approach $K^2$ times fewer computations than standard convolution. For ParConv, under the reasonable assumption that $C_{\rm{in}}=C_{\rm{out}}$ and $C_{\rm{out}}>>K^2$, the FLOPs ratio to standard convolution can be rewritten as follows:


\begin{eqnarray}
\frac{1}{{2{K^2}}} + \frac{{3\omega }}{{4{K^2}}}
\end{eqnarray}

Obviously, the computational cost can be adjusted by changing the value of the channel multiplier $\omega$.

\begin{table}
\caption{The FLOPs ratios of compact convolutions to standard convolution}
\centering \label{RadioFLOPs}
\begin{tabular}{|c|c|}
\hline
Type &  FLOPs Ratio   \\
\hline
DSConv      &  $\frac{1}{C_{\rm{out}}}+\frac{1}{K^{2}}$   \\
\hline
ParConv       &   $\frac{1}{2K^2}+\frac{\omega}{2}(\frac{1}{K^2}+\frac{1}{C_{\rm{out}}}+\frac{C_{\rm{in}}}{2{C_{\rm{out}}}K^2})$         \\
\hline
\end{tabular}
\end{table}

\subsection{Knowledge Distillation with Multiple Losses}
In order to reduce the performance gap between the standard CNN and the corresponding ParConv-based compact CNN (CCNN), knowledge distillation is necessary. Three kinds of training losses are included in the process of knowledge distillation, i.e., Kullback-Leibler (KL) divergence loss, cross entropy (CE) loss and solving procedure (SP) loss. The final loss is formulated as the weighted sum of these losses:

\begin{eqnarray}
\label{loss}
l = \mu  {l_{{\rm{KL}}}} + \beta {l_{{\rm{CE}}}} + \lambda {l_{\rm{SP}}}
\end{eqnarray}

The CE loss with so-called hard labels is the most common training criterion in classification tasks and is defined as follows:

\begin{eqnarray}
\label{ce}
{l_{{\rm{CE}}}} = -\sum\limits_t {\log p({s_t}|{{\bf{x}}_t})}
\end{eqnarray}
where ${\log p({s_t}|{{\bf{x}}_t})}$ is the estimated posterior probability of the target class $s_t$ from the CNN output given the input $\boldsymbol{x}_t$.

The KL divergence is a measure of how one probability distribution is different from another probability distribution. In our approach, it is used to compute the difference between the output distribution of standard CNN ${p_{\rm{S}}}(s|{{\boldsymbol{x}}_t})$ and the corresponding distribution from CCNN ${p_{\rm{C}}}(s|{{\boldsymbol{x}}_t})$:

\begin{eqnarray}
\label{kl1}
\begin{split}
{l_{{\rm{KL}}}} =& \sum\limits_t {\sum\limits_s {{p_{\rm{S}}}(s|{{\bf{x}}_t})} } \log (\frac{{{p_{\rm{S}}}(s|{{\bf{x}}_t})}}{{{p_{\rm{C}}}(s|{{\bf{x}}_t})}})\\
  =& \sum\limits_t {\sum\limits_s {[{p_{\rm{S}}}(s|{{\bf{x}}_t})\log {p_{\rm{S}}}(s|{{\bf{x}}_t}) }} \\
   & { {- {p_{\rm{S}}}(s|{{\bf{x}}_t})\log {p_{\rm{C}}}(s|{{\bf{x}}_t})]} }
\end{split}
\end{eqnarray}

Because we only optimize the CCNN, the KL loss in Eq.~(\ref{kl1}) can be rewritten to retain:
\begin{eqnarray}
\label{kl2}
{l_{{\rm{KL}}}} =  - \sum\limits_t {\sum\limits_s {{p_{\rm{S}}}(s|{{\bf{x}}_t})\log {p_{\rm{C}}}(s|{{\bf{x}}_t})} }
\end{eqnarray}

Essentially, the KL loss in Eq. (\ref{kl2}) is simplified to CE loss with soft labels. The weighted sum of KL loss and CE loss is:

\begin{eqnarray}
\label{klandce}
\begin{split}
\mu {l_{{\rm{KL}}}} + \beta {l_{{\rm{CE}}}} =& - \sum\limits_t {\sum\limits_s {\mu {p_{\rm{S}}}} } (s|{{\boldsymbol{x}}_t})\log {p_{\rm{C}}}(s|{{\boldsymbol{x}}_t}) \\
&  - \sum\limits_t {\beta \log {p_{\rm{C}}}({s_t}|{{\bf{x}}_t})} \\
 =&  - \sum\limits_t {[(\beta  + \mu {p_{\rm{S}}}({s_t}|\boldsymbol{x_t}))\log {p_{\rm{C}}}({s_t}|{{\boldsymbol{x}}_t}) }\\
 & {+ \sum\limits_{s \ne {s_t}} \mu  {p_{\rm{S}}}(s|\boldsymbol{x_t})\log {p_{\rm{C}}}(s|{{\boldsymbol{x}}_t})]}
\end{split}
\end{eqnarray}
From the above formula, it is clear that the CE helps the model focus on the important parts by providing prior knowledge (ground truth of inputs).

Furthermore, it is not enough to give the CCNN the answers to problems from the standard CNN. A better teacher always explains the solving procedures of problems so that the students can handle such problems from learning one certain example. In CNN, we define a series of solving procedure matrices (SPMs). An SPM is the result of elementwise subtraction between the extracted attention feature maps from two layers, which is intuitively reasonable by using the change in outputs between two layers to represent the solving procedure. The attention map in a layer needs to emphasize valuable information for the following flow. Naturally, assuming a convolutional layer has the output tensor ${\boldsymbol{O}} \in {R^{D \times D \times C}}$ with each feature map ${{\boldsymbol{O}}_c} \in {R^{D \times D}} $, the attention map can be simply computed as \cite{zagoruyko2016paying}:
\begin{eqnarray}
\label{AM}
\boldsymbol{A} = \sum\limits_{c = 1}^C {{\boldsymbol{O}_c}}
\end{eqnarray}

The SPM $\boldsymbol{S}$ for the $i$-th layer and the $j$-th layer ($j>i$) is:

\begin{eqnarray}
\label{SPM}
\boldsymbol{S} = \boldsymbol{A}_j - \boldsymbol{A}_i
\end{eqnarray}

Finally, the SP loss can be obtained as follows:

\begin{eqnarray}
\label{SP}
{l_{{\rm{SP}}}} = \sum\limits_t {\sum\limits_{n = 1}^N {\frac{1}{N} \times \left\| {\frac{{{{\boldsymbol{S}}_{{{\rm{S}}_n}}}({{\boldsymbol{x}}_t})}}{{{{\left\| {{{\boldsymbol{S}}_{{{\rm{S}}_n}}}({{\boldsymbol{x}}_t})} \right\|}_{\rm{F}}}}} - \frac{{{{\boldsymbol{S}}_{{{\rm{C}}_n}}}({{\boldsymbol{x}}_t})}}{{{{\left\| {{{\boldsymbol{S}}_{{{\rm{C}}_n}}}({{\boldsymbol{x}}_t})} \right\|}_{\rm{F}}}}}} \right\|} } _{\rm{F}}^2
\end{eqnarray}

\noindent where $N$ is the total number of SPMs in the CNN, ${{\boldsymbol{S}}_{{{\rm{S}}_n}}}$ is the $n$-th SPM in the standard CNN, ${{\boldsymbol{S}}_{{{\rm{C}}_n}}}$ is the corresponding SPM in the ParConv-based CCNN, and ${\left\|  \bullet  \right\|_F}$ is the standard Frobenius norm.

\section{Experiments}
\label{sec:exp}
The proposed distillation algorithm is mainly validated on offline handwritten Chinese text recognition (HCTR) using the CASIA database \cite{liu2011casia,yin2013icdar}. In addition, in order to accurately observe the performance changes of CNNs, a 5-gram LM \cite{katz1987estimation} is only added in our final results. PyTorch \cite{paszke2017automatic} is used as a deep learning platform in all experiments.





\subsection{DCNN on CASIA}
The baseline DCNN architecture in \cite{wang2019writer} is adopted. Please note that except for the categories of output layers, the CNNs in \cite{wang2018comprehensive,wang2019writer} have the same architecture. According to the configuration in \cite{wang2019writer}, both offline isolated handwritten Chinese character datasets (HWDB1.0, HWDB1.1 and HWDB 1.2) and the training sets of offline handwritten Chinese text datasets (HWDB2.0, HWDB2.1 and HWDB2.2) are used. In total, there are 7,360 classes (Chinese characters, symbols, garbage) and 3,932,197 images. After extracting frame-level images from the original datasets, there are 148,648,249 training samples for the training of DCNN. The ICDAR 2013 competition set is adopted as the evaluation set \cite{yin2013icdar}. The CER is computed as:
\begin{equation}
\text{CER} = \frac{{{N_\text{s}} + {N_\text{i}} + {N_\text{d}}}}{N}
\end{equation}
where $N$ is the total number of character samples in the evaluation set. $N_\text{s}$, $N_\text{i}$ and $N_\text{d}$ denote the number of substitution errors, insertion errors and deletion errors, respectively. In this study, we do not use additional language models because we focus on the performance of the CNN.

Each class is modeled by 3 tied HMM states on average. The input of DCNN is a normalized frame-level image of size 40$\times$40 extracted from original images, and then each frame is extended to 48$\times$48 by adding the margin. The output layer has 22,080 (7,360$\times$3) output nodes. In the DCNN architecture, there are 14 convolutional layers that use standard 2D convolution and are followed by batch normalization (BN) and nonlinearity activation ReLU. The number of channels continuously increases from 100 to 700. Except for the first and last convolutional layers without padding, other convolutional layers have the same padding value of 1. The stride is set to 1 for all convolutional layers, while the stride of all max pooling layers is 2 with a 3$\times$3 window.

According to our proposed guideline in Algorithm~\ref{alg:algorithm_dis_train}, we first analyze the computational costs and storage of convolutional layers (Convs) and fully-connected layers (FCs) in DCNN. The details of the DCNN and statistical results are shown in Table~\ref{DCNNAnalysis}. Here, we do not consider the consumption of the batch normalization (BN) operation and max-pooling (MaxPooling) operation because they occupy a negligible part. Based on the analyzed results, it is necessary to reduce the parameters in fully-connected layers due to a large proportion of the storage (35\%). In addition, we can observe that almost all computational costs are generated by convolutional layers.

\begin{table}
\caption{Architecture and quantitative analysis of DCNN character model. The abbreviations f, k, s, p represent the number of feature maps, kernel size, stride length and padding size, respectively.}
\centering \label{DCNNAnalysis}
\scalebox{0.75}{
\begin{tabular}{|c|c|c|c|c|c|c|}
\hline
Layer & Configurations  &   Spatial Size  & FLOPs ($\times 10^8$) & Fraction  &  Storage (MB) & Fraction        \\
\hline
FC2  & 500$\times$22080 &  1$\times$1   & 0.1104  & 0.69\% & 42.1985 & 33.90\%  \\ 
FC1  & 700$\times$500 &    1$\times$1     &0.0035 & 0.02\% & 1.3371 & 1.07\% \\
\hline
Conv5 & F:700, K:1$\times$1, S:1, P:0 & 1$\times$1     &  0.0049  & 0.03\% & 1.8826 & 1.51\%  \\  
\hline
MaxPooling & K:3$\times$3, S:2 &-  &  - & - & - & - \\
\hline
Conv4\_4 & F:700, K:3$\times$3, S:1, P:1 & 4$\times$4 &0.7056 & 4.40\% & 16.8362 & 13.52\% \\ 
Conv4\_3 & F:700, K:3$\times$3, S:1, P:1 &4$\times$4 &0.6048 & 3.78\% & 14.4329 & 11.59\% \\ 
Conv4\_2 & F:600, K:3$\times$3, S:1, P:1 &4$\times$4 &0.4320 & 2.70\% & 10.3111 & 8.28\% \\ 
Conv4\_1 & F:500, K:3$\times$3, S:1, P:1 & 4$\times$4 & 0.3600 & 2.25\% & 8.5926 &  6.90\% \\ 
\hline
MaxPooling & K:3$\times$3, S:2 & -& - & - & - & - \\
\hline
Conv3\_4 & F:500, K:3$\times$3, S:1, P:1 &10$\times$10& 2.2500 & 14.04\%& 8.5926 & 6.90\% \\ 
Conv3\_3 & F:500, K:3$\times$3, S:1, P:1 &10$\times$10 &1.8000 & 11.23\%& 6.8760 & 5.52\% \\
Conv3\_2 & F:400, K:3$\times$3, S:1, P:1 &10$\times$10& 1.0800 & 6.74\% & 4.1275 & 3.32\% \\ 
Conv3\_1 & F:300, K:3$\times$3, S:1, P:1 &10$\times$10  &0.8100 & 5.06\% & 3.0956 & 2.49\% \\
\hline
MaxPooling & K:3$\times$3,S:2 & - &- & - & - & - \\
\hline
Conv2\_4 & F:300, K:3$\times$3, S:1, P:1 &    22$\times$22 &3.9204  & 24.47\%& 3.0956 & 2.49\% \\
Conv2\_3 & F:300, K:3$\times$3, S:1, P:1 &  22$\times$22   &2.6136  & 16.31\%& 2.0657 & 1.66\% \\
Conv2\_2 & F:200, K:3$\times$3, S:1, P:1 &   22$\times$22  & 0.8712  & 5.44\% & 0.6905 & 0.55\% \\
Conv2\_1 & F:100, K:3$\times$3, S:1, P:1 &  22$\times$22   & 0.4356  & 2.72\% & 0.3452  & 0.28\% \\
\hline
MaxPooling & K:3$\times$3, S:2 & - &- & - & - & - \\
\hline
Conv1 & F:100, K:3$\times$3, S:1, P:0 &  46$\times$46  &  0.0190 & 0.12\% & 0.0053 & 0.00\% \\
\hline
Input &  Frame-level image    &  48$\times$48  &   - & - & - & -  \\
\hline
\end{tabular}}
\end{table}

\subsection{Experiments on Architecture Distillation}
In order to make a fair comparison, the setting of hyperparameters in the training stage is the same for all experiments in this part. The minibatch size is 1,000 in each iteration, the momentum is 0.9 and the weight decay is 0.0001. The learning rate is initially set to 0.1 and decreased by 0.92 after every 4,000 iterations. After two epochs are conducted, the learning rate is reduced to 0.0002 and the networks are convergent. For fully-connected layers, although many algorithms can be used, we choose the simplest algorithm that controls the weights of fully-connected layers by adjusting the feature dimension before the output layer. The effectiveness of this strategy has also been reported in \cite{xu2019lightweightnet}. In Table~\ref{RadioFC}, different bottleneck features are compared. The 500 dimension corresponds to the baseline DCNN. Three observations can be found. First, because the FLOPs of fully-connected layers occupy a small proportion (less than 1\%), the total FLOPs remains almost unchanged when the dimension is below 100. Second, with the dimension changing from 500 to 50, the storage decreases considerably and the CER has fluctuates slightly, which means that most parameters in fully-connected layers are redundant and can be safely ignored \cite{denil2013predicting}. Finally, when the dimension is smaller than 50, the storage tends to be stable, as most storage consumption is caused by convolutional layers, however, it is reasonable to observe that the CER begins to increase due to the very small number of parameters in the bottleneck feature leading to missing information. Considering the tradeoff between storage ratio $\gamma$ about fully-connect layers (see Algorithm~\ref{alg:algorithm_dis_train}) and CER, we choose dimension 50 in the following experiments.

\begin{table}
\caption{FLOPs ($\times10^8$), STORAGE (MB), RATIO $\gamma (\%)$  and CER (\%) comparison by using the different bottleneck features in fully-connected layers.}
\centering \label{RadioFC}
\scalebox{0.9}{
\begin{tabular}{|c|c|c|c|c|}
\hline
Low-dimensional Feature &  FLOPs ($\times 10^8$)    &  Storage (MB) & $\gamma (\%)$   & CER (\%)  \\
\hline
500      &   16.02     & 124.5 & 34.97  &  9.17 \\
\hline
100      &   15.96       &     89.74  &   9.78  &  9.04      \\
\hline
50     &  15.92        &     85.39   &  5.19 &  9.01     \\
\hline
25     &  15.91      &     83.22  &  2.71 &  9.38     \\
\hline
20     &  15.91        &     82.79  & 2.20     &  9.40         \\
\hline
\end{tabular}}
\end{table}

Based on the DCNN with bottleneck feature 50 (DCNN\_LF50), except for the initial and last convolutional layers, we replace all remaining 12 standard convolutional layers with our ParConv blocks (same channel multiplier $\omega$ for all 12 layers). The reconstructed compact CNN is notated as ParCNN\_$\omega$. For example, the ParCNN\_$\omega0.5$ indicates that the value of $\omega$ in all ParConv blocks is set to 0.5. In order to demonstrate that the proposed ParConv is a more efficient and effective replacement for standard convolution, we compare it with depthwise separable convolution (DSConv) and the architecture distillation algorithm LightweightNet proposed in \cite{xu2019lightweightnet}. Besides, we also construct the simplified ParConv (SParConv) by removing the pointwise convolution before depthwise separable convolution block in ParConv to verify the role of pointwise convolution. We directly adopt DSConv and SParConv to replace the same 12 standard convolutional layers and build the corresponding compact CNN: DSCNN and SParCNN, respectively. Table~\ref{RadioConv} lists all related results. The notation *\_Res indicates that we add residual connections for all corresponding compact blocks, namely, there is another path directly connecting the input and output of the compact block.

\begin{table}
\caption{The detailed results of using different compact blocks to directly replace standard convolution without knowledge distillation.}
\centering \label{RadioConv}
\begin{tabular}{|c|c|c|c|}
\hline
Model &  FLOPs ($\times 10^8$) &  Storage (MB)    & CER (\%)   \\
\hline
DCNN\_LF50      &   15.92       & 89.74           &  9.01     \\
\hline
DSCNN \cite{chollet2017xception} &    1.84       &    15.46        &  19.44     \\
DSCNN\_Res  &    2.67        &       19.79         &  10.01 \\
\hline
LightweightNet \cite{xu2019lightweightnet}        &  2.12        &     23.41        &  10.30     \\
\hline
SParCNN    & 1.81  & 15.30 & 10.45  \\
\hline
ParCNN\_$\omega$0.5 &    1.56      &      14.14       &   10.44   \\
ParCNN\_$\omega$0.5\_Res & 2.38  & 18.46  &  10.03 \\
\hline
ParCNN\_$\omega$1 &  2.21        &     17.41        &  10.00      \\
ParCNN\_$\omega$1\_Res & 3.03 &  21.74 &  9.80 \\
\hline
ParCNN\_$\omega$2 &     3.50    &   23.95        & 9.72        \\
ParCNN\_$\omega2$\_Res &   4.32   &   28.29  &  9.54 \\
\hline
ParCNN\_$\omega$4 &   6.07      &   37.04         &   9.59     \\
ParCNN\_$\omega4$\_Res & 6.90 &  41.37 &  9.53 \\
\hline
\end{tabular}
\end{table}

\begin{itemize}
\item Comparison with different values of $\omega$
\end{itemize}

We changed the value of $\omega$ from 0.5 to 4. As shown in Table~\ref{RadioConv}, the CERs of ParCNN and ParCNN\_Res consistently decrease from 10.44\% to 9.59\% and 10.03\% to 9.53\%, respectively. Naturally, the computational resources also increase with the increment of $\omega$. Compared with the network DCNN\_LF50, the ParCNN without residual connection can achieve 10.21$\times$ to 2.62$\times$ FLOPs based improvement and 6.35$\times$  to 2.42$\times$ in storage reduction, while the corresponding ParCNN\_Res can reduce FLOPs from 6.69$\times$ to 2.31$\times$  and storage overhead from 4.86$\times$ to 2.17$\times$. By comparing the results of ParCNN and ParCNN\_Res, we can observe that the residual connection always yields a performance improvement. Additionally, the residual connection introduces extra computational resources (approximately 0.82$\times10^8$ FLOPs and 4.33MB) due to the pointwise convolution necessary for the special situation where the number of input and output channels of a compact block is different. Actually, in most other typical CNNs where the number of input and output channels is the same for most convolutional layers, the residual connection does not lead to too much extra consumption.

\begin{itemize}
\item Comparison with depthwise separable convolution
\end{itemize}

First, from the DSCNN and DSCNN\_Res results, an interesting and reasonable observation can be made. Without the residual connection, the recognition performance of DSCNN significantly declines with a CER of 19.44\%. In essence, directly using DSConv to replace standard convolution doubles the depth of the network, which easily leads to the degradation of network \cite{he2016deep}. After adding the residual connection, the recognition performance of DSCNN\_Res returns to a normal value (10.07\%). This phenomenon reflects another advantage of our ParConv, i.e., relaxing the residual connection requirement, which can reduce the possibility of additional computations. In the case of similar CERs (10.0\%), the proposed ParConv-based compact CNNs (ParCNN\_$\omega$0.5\_Res, ParCNN\_$\omega$1) consume significantly fewer computing resources (2.38$\times10^8$, 2.21$\times10^8$ vs. 2.67$\times10^8$ in FLOPs and 18.46MB, 17.41MB vs. 19.79MB in storage).


\begin{itemize}
\item Comparison with simplified parsimonious convolution
\end{itemize}

 In SParConv, a quarter of the input channels are fed into depthwise separable convolution. Compared with the proposed ParCNN, all the performance indicators of SParCNN are worse than those of the proposed ParCNN\_0.5, which indicates the importance of the pointwise convolution in the ParConv. Besides, the parameter $\omega$ in ParConv can be set to different values so that we can choose a compression rate that meets the performance requirements.

\begin{itemize}
\item Comparison with LightweightNet
\end{itemize}

We also reproduce the architecture distillation algorithm \cite{xu2019lightweightnet}.
The reconstructed LightweightNet needs 2.12$\times10^8$ FLOPs and occupies 23.41MB while the corresponding CER is 10.30\%. The network ParCNN\_$\omega$0.5 with a comparable recognition performance can apparently outperform it in FLOPs and storage. Meanwhile, the networks ParCNN\_$\omega$0.5\_Res and ParCNN\_$\omega$1 have similar FLOPs with LightweightNet, but lower storage and CERs.

\subsection{Experiments on Architecture and Knowledge Distillation}
As shown in Table~\ref{RadioConv}, a smaller value of $\omega$ can obtain a larger compression ratio but suffer from worse recognition performance. In order to reduce the performance gap between the baseline CNN and the compact CNN, it is necessary to introduce knowledge distillation. In knowledge distillation, i.e., based on Eq. (\ref{loss}), the weight $\mu$ is set to 0.8, $\beta$ equals 0.2 and $\lambda$ is 0.1. Except for the batch size set to 700, all other initial training hyper parameters are the same as the parameters in architecture distillation.

In order to excavate the best capability of the proposed approach, we first combine knowledge distillation to improve the recognition performance of the smallest network ParCNN\_$\omega$0.5. From the results of Table~\ref{ArcandKnowConv}, we can observe that the knowledge distillation can yield remarkable reductions of CER: from 10.44\% to 9.79\% (+KL Loss), 9.94\% (+SP Loss) and 9.68\% (+KL Loss \& +SP Loss), which demonstrates the effectiveness of the SP loss and the necessity of knowledge distillation. Compared with the baseline DCNN, our proposed joint architecture and knowledge distillation can achieve a 10$\times$ reduction in computational cost and 9$\times$ storage compression with only a 0.51\% increment in CER, i.e., a relative CER increment of 5.6\%.

\begin{table}
\caption{The results of joint architecture and knowledge distillation for DCNN.}
\centering \label{ArcandKnowConv}
\begin{tabular}{|c|c|c|c|}
\hline
Model &  FLOPs ($\times 10^8$) &  Storage (MB)    & CER (\%)   \\
\hline
DCNN                                     &   16.02     & 124.5   &  9.17 \\
\hline
ParCNN\_$\omega$0.5 &  \multirow{4}{*}{ 1.56}       &      \multirow{4}{*}{14.14}       &   10.44   \\
+KL                 &                               &                                   &   9.79 \\
+SP                &                                &                                   &   9.94 \\
+KL+SP             &                               &                                   &   9.68\\
\hline

\end{tabular}
\end{table}

Table~\ref{ArcandKnoDisforDiffNet} shows the results for different acceleration and compression ratios based on channel multiplier $\omega$. It can be observed that the value of $\omega$ can effectively control the acceleration and compression ratio and recognition performance. When $\omega$ is set to 1, compared with the baseline DCNN, the proposed approach can reduce the computational cost and model size by $>$7$\times$ with a relative CER increment of 2.2\%. If we further increase the value of $\omega$ to 2, the compact network ParCNN\_$\omega$2 with a $>$4$\times$ acceleration ratio and $>$ a 5$\times$ compression ratio can even obtain a better performance (9.09\% vs. 9.17\%).

\begin{table}
\caption{The results of the proposed approach for different acceleration and compression ratios based on channel multiplier $\omega$.}
\centering \label{ArcandKnoDisforDiffNet}
\scalebox{0.9}{
\begin{tabular}{|c|c|c|c|c|}
\hline
Model &  FLOPs ($\times 10^8$) &  Storage (MB)    & Without KD  & With KD    \\
\hline
DCNN                                     &   16.02     & 124.5   &  9.17 & - \\
\hline
ParCNN\_$\omega$0.5 &  1.56       &     14.14       &   10.44 &  9.68 \\
\hline
ParCNN\_$\omega$1 &  2.21       &     17.41       &   10.00  & 9.37 \\
\hline
ParCNN\_$\omega$2 &  3.50       &     23.95       &   9.72  & 9.09 \\
 \hline
\end{tabular}}
\end{table}

\begin{figure}
\centering
\subfigure[The comparison of KL loss.]{
\label{fig:subfig:c} 
\includegraphics[width=3.5in]{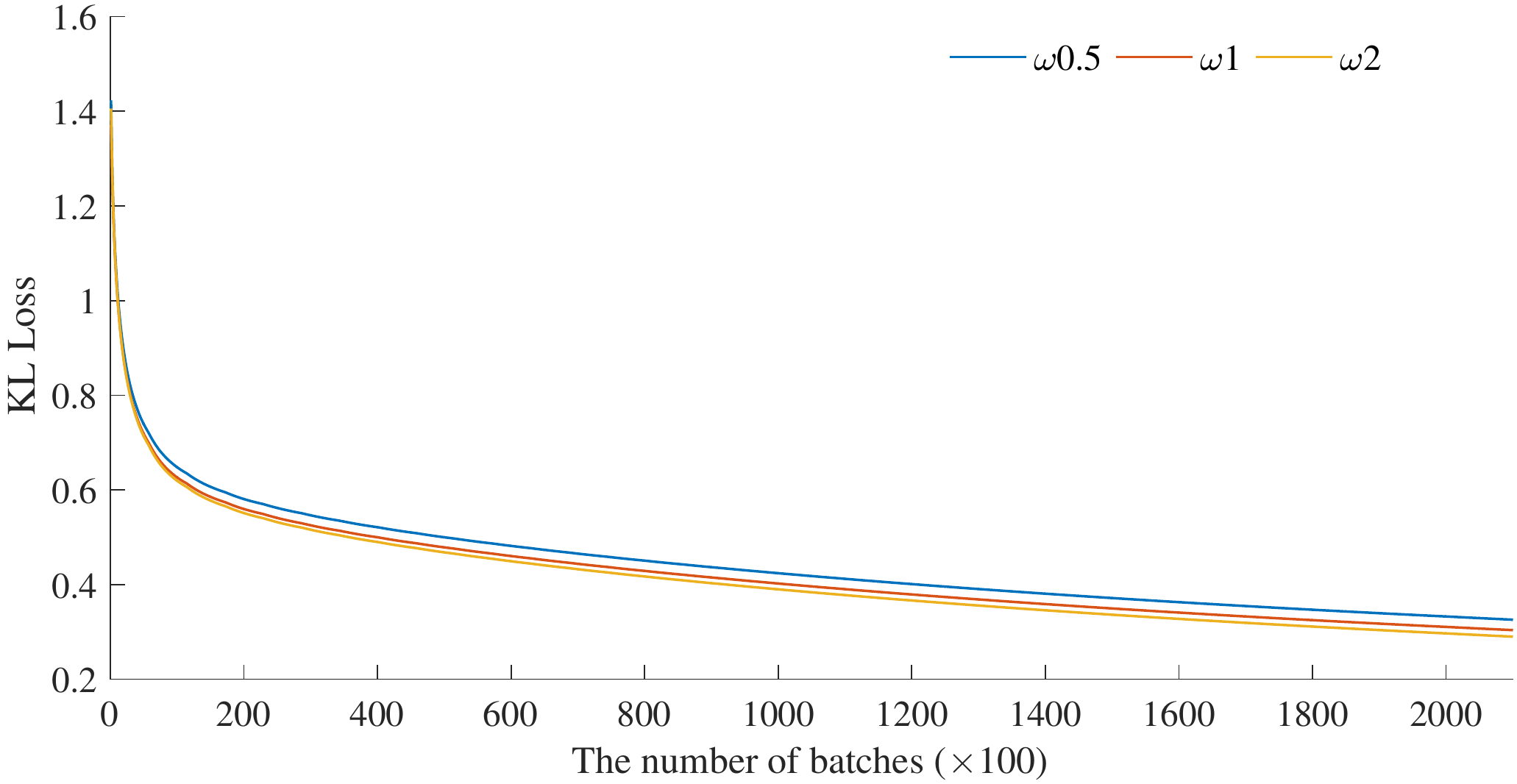}}

\subfigure[The comparison of CE loss.]{
\label{fig:subfig:d} 
\includegraphics[width=3.5in]{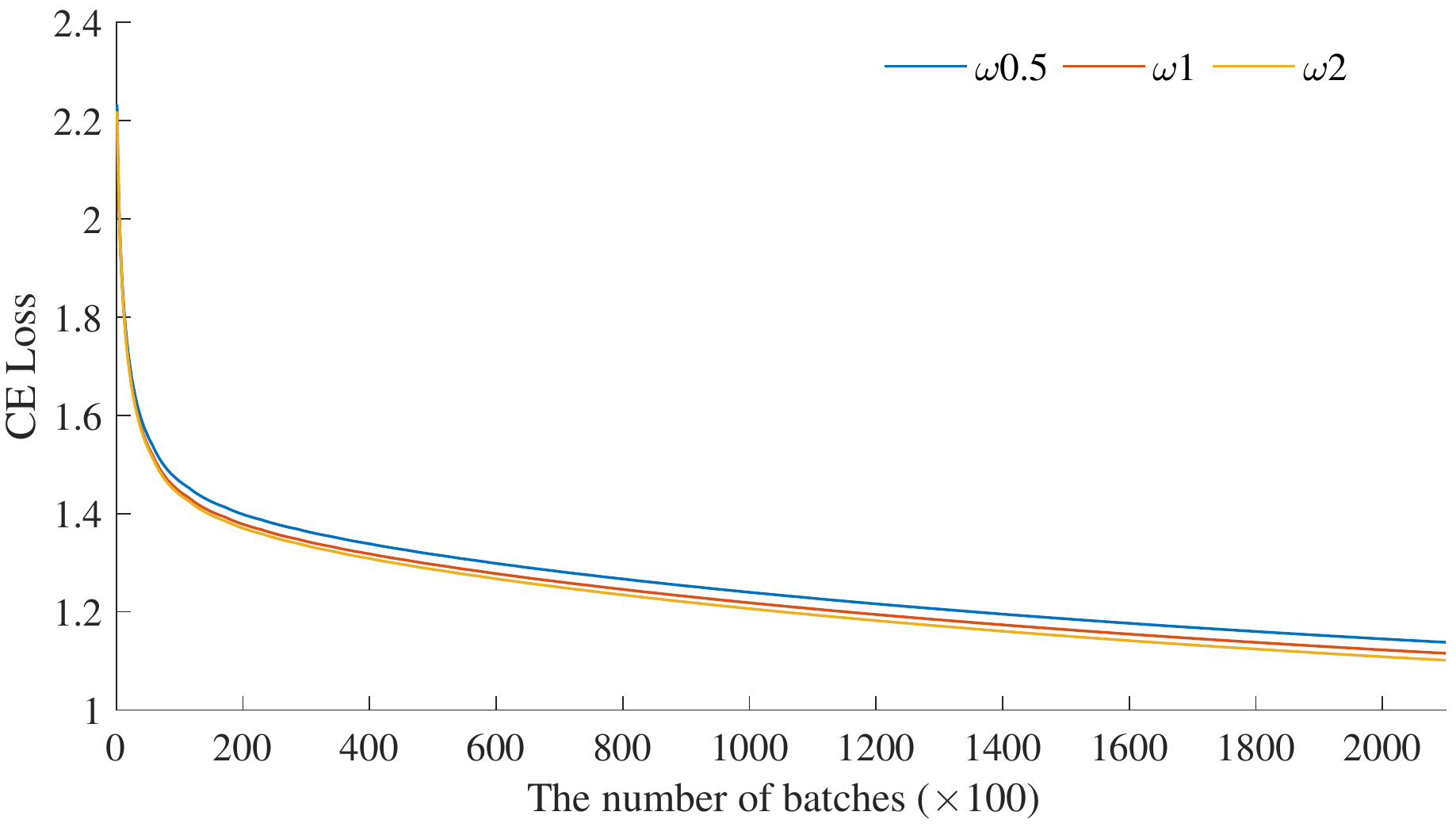}}

\subfigure[The comparison of SP loss.]{
\label{fig:subfig:e} 
\includegraphics[width=3.5in]{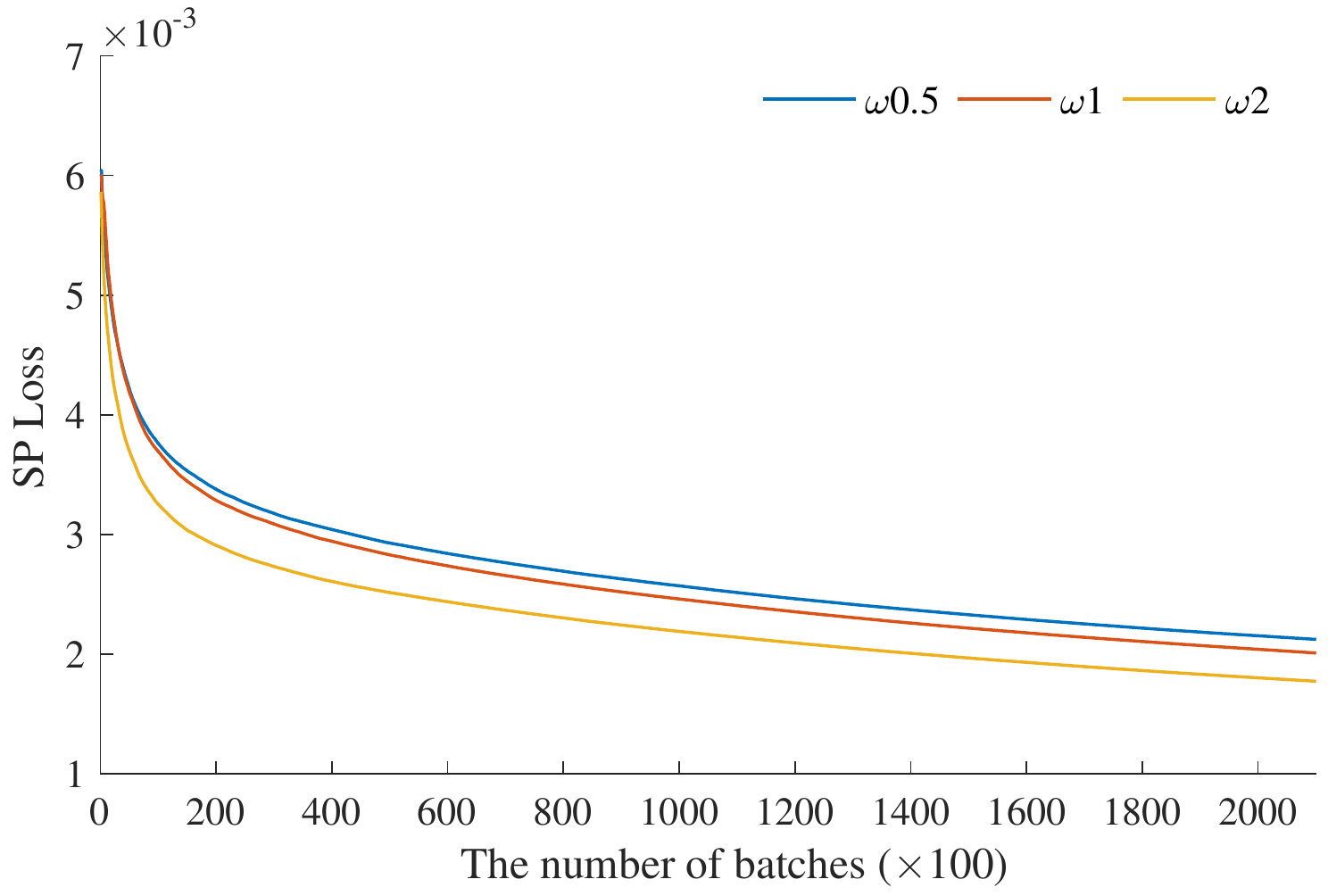}}
\caption{The comparison of multiple losses for ParCNN\_$\omega$0.5, ParCNN\_$\omega$1 and ParCNN\_$\omega$2 when all losses are considered simultaneously in the training stage. For simplicity, we use $\omega$0.5, $\omega$1, $\omega$2 to represent respective networks in all figures. }
\label{trainingloss}
\end{figure}

In order to better understand why more parameters can yield better performance, we draw the learning curves of multiple losses during training for ParCNN\_$\omega$0.5, ParCNN\_$\omega$1 and ParCNN\_$\omega$2 in Fig.~\ref{trainingloss}. It can be observed that all kinds of losses decrease with increasing $\omega$, which is in line with our expectations. Another interesting observation is that the relative gap of SP loss between ParCNN\_$\omega$1 and ParCNN\_$\omega$2 is larger than other kinds of losses. This indicates that SP loss should play an important role in the training of ParCNN\_$\omega$2.

Finally, considering that LM plays an important role in HCTR, we add the same 5-gram LM \cite{wang2019writer} to compare the final results of DCNN and ParCNN\_$\omega$0.5. As shown in Table~\ref{Finalresults}, it is reasonable to observe that the performance gap is almost fixed by LM, which indicates that the proposed algorithm can yield a remarkable compression ratio with negligible accuracy loss. We also test the actual runtime (milliseconds per batch) for DCNN and the proposed ParCNN\_$\omega$0.5. All models with batch size 120 are run 10 times in the same machine that is equipped with PyTorch (version 1.0.1) with GeForce RTX 2080, CUDA version 10.0.130 and CUDNN \cite{chetlur2014cudnn} version 7402. Although FLOPs reduction (theoretical) is amazing, the practical speedup (2$\times$) is limited. The main reason is that the 1$\times$1 convolutions and depthwise convolutions in PyTorch are relatively slow, and the latest CUDNN library is specially optimized for 3$\times$3 convolutions. However, we observe that when running the same batch, the DCNN consumes 3,353 MB of GPU memory while the ParCNN\_$\omega$0.5 only needs 759 MB.

\begin{table}
\caption{The comparison of final results after adding the same 5-gram LM.}
\centering \label{Finalresults}
\begin{tabular}{|c|c|c|c|c|c|}
\hline
Model &  FLOPs ($\times 10^8$) &  Storage (MB)  & GPU Time (ms/batch) & GPU Occupancy (MB)   & CER (\%)   \\
\hline
DCNN \cite{wang2019writer} & 16.02              & 124.5     & 39.3       &  3,353   &  3.52 \\
\hline
ParCNN\_$\omega$0.5     &  1.56      &  14.14    &    19.0  & 759 &  3.55  \\
\hline
\end{tabular}
\end{table}

\begin{figure}
\centering
\subfigure[The weight distribution of DCNN.]{
\label{fig:subfig:a} 
\includegraphics[width=2.9in,height=1.7in]{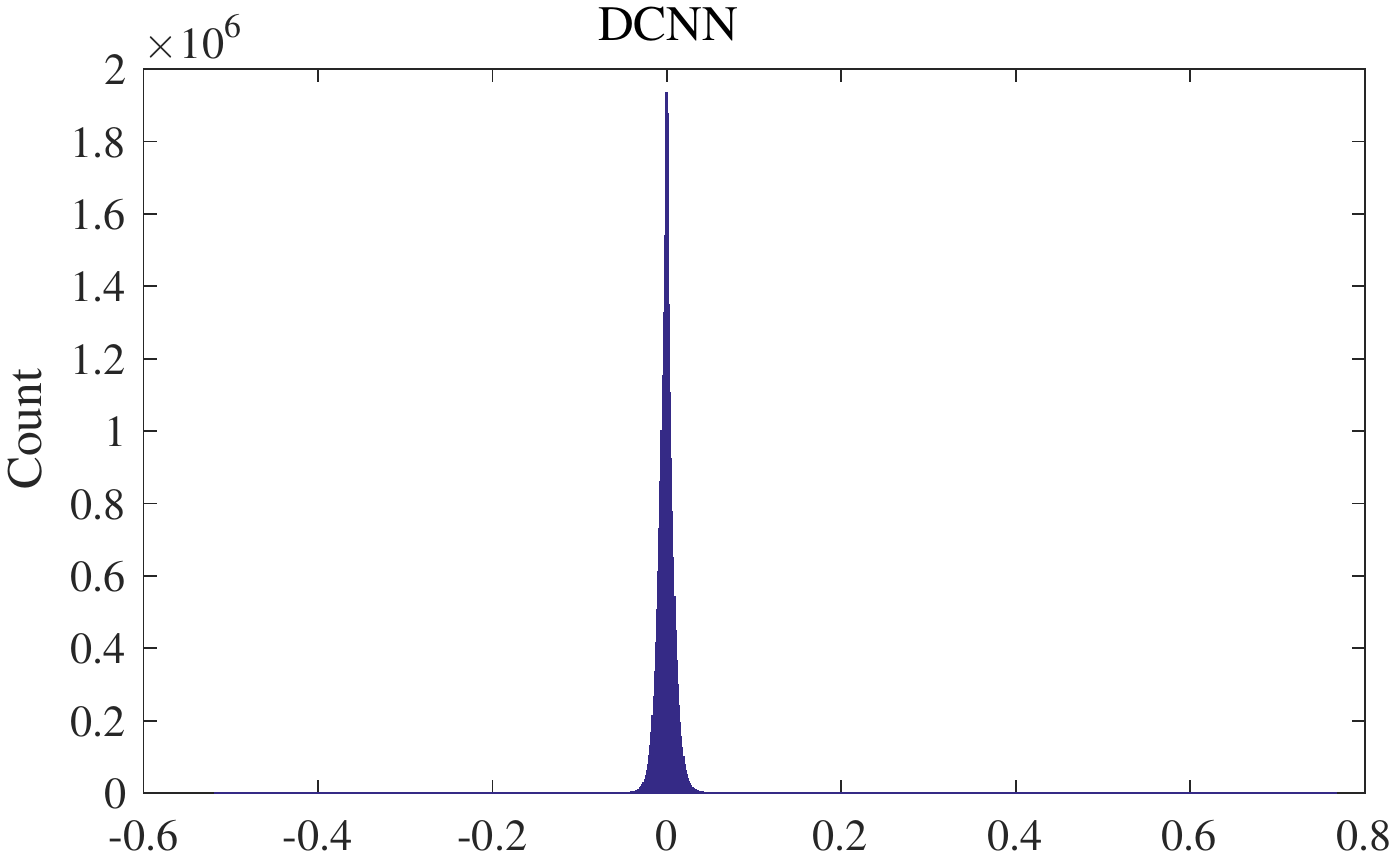}}
\subfigure[The weight distribution of ParCNN\_$\omega$0.5.]{
\label{fig:subfig:b} 
\includegraphics[width=2.9in,height=1.7in]{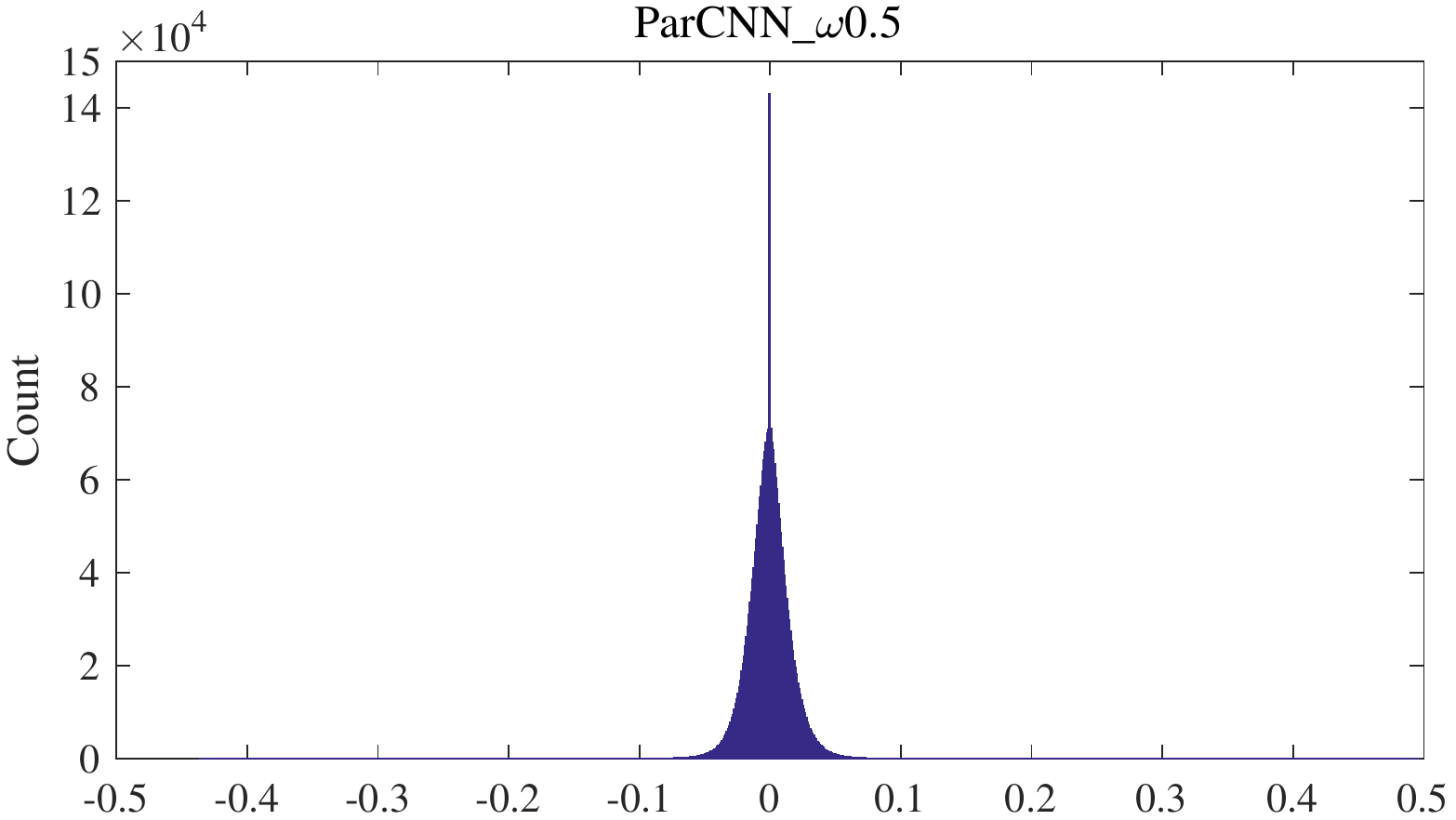}}
\caption{Weight analysis for DCNN and ParCNN\_$\omega$0.5.}
\label{weightana}
\end{figure}

Furthermore, in Fig.~\ref{weightana}, it is obvious that the weights of DCNN are more concentrated around 0, which indicates massive weights in DCNN may be unimportant. The utilization of weights greatly improves in ParCNN\_$\omega$0.5, where the weight distribution becomes flatter. To some extent, this can explain why a small network can also achieve similar recognition accuracy. In order to demonstrate that the proposed approach can also be successfully applied on mainstream backbone networks, in the following experiments, we reconstruct the corresponding compact networks according to the structures of Res50 and Res18 and conduct experiments on CTW and MNIST respectively.

\subsection{{Experiments on CTW}}

\begin{figure*}
\centering
\includegraphics[width=3.5in,height=2.5in]{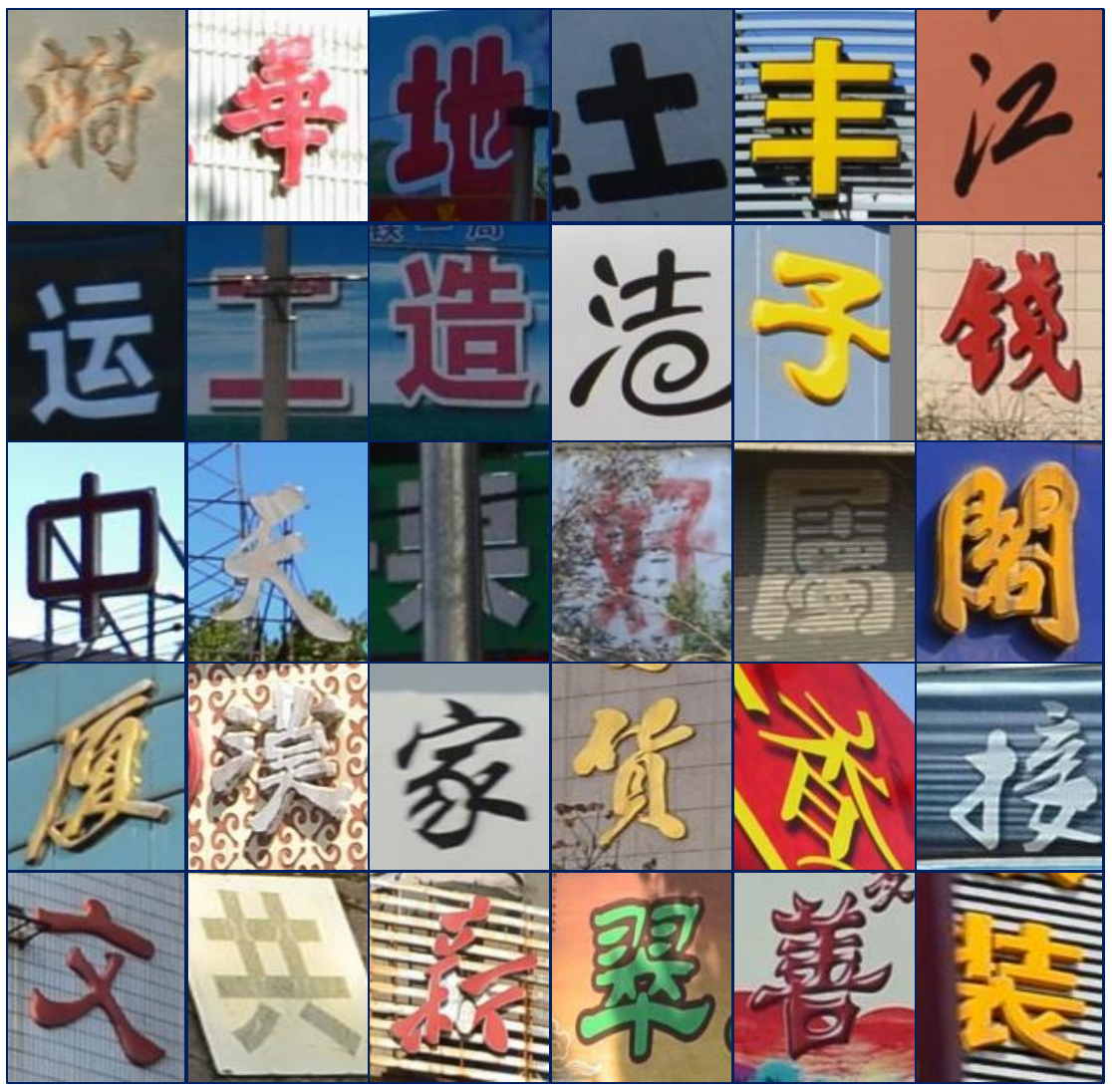}
\caption{Some examples in the CTW dataset.}
\label{CTWData}
\end{figure*}

\begin{figure*}
\centering
\includegraphics[width=4.5in,height=1.5in]{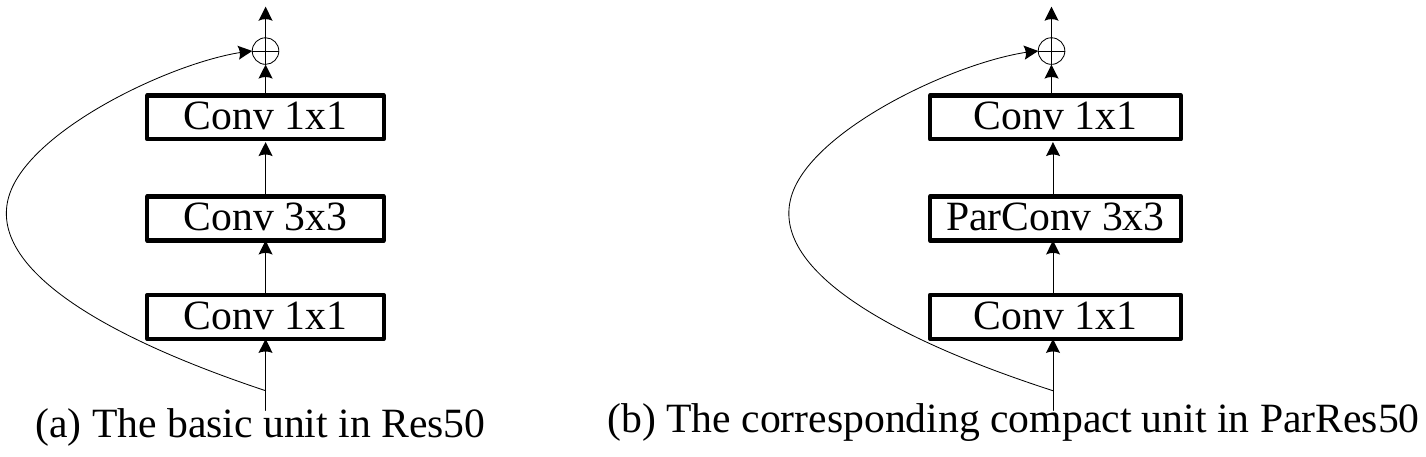}
\caption{The comparison of basic units in Res50 and ParRes50.}
\label{Res50ParRes50}
\end{figure*}

The CTW dataset contains 1,019,402 Chinese character images extracted from 32,285 street view images. The number of Chinese character categories is 3,850. These character images are annotated to different attributes: occlusion, complex background, distortion, raised character, word art and handwritten character. Examples are illustrated in Fig.~\ref{CTWData}. The image preprocessing we use is approximately consistent with the method in \cite{yuan2018chinese}. We first train a Res50 \cite{he2016deep} network. As in \cite{yuan2018chinese}, only the top 1,000 frequent Chinese character categories are considered. As shown in Fig.~\ref{Res50ParRes50}, the basic unit of Res50 includes a 1$\times$1 convolutional layer followed by a 3$\times$3 convolutional layer, and a 1$\times$1 convolutional layer in the end. The first 1$\times$1 convolutional layer can form a bottleneck to reduce the total parameters. We can easily build the corresponding compact network ParRes50 by replacing the standard 3$\times$3 convolution with the proposed ParConv with channel multiplier $\omega = 0.5$ (see Fig.~\ref{Conv}). In the training of Res50 and ParRes50, the minibatch size is 64, and the momentum is 0.9. The learning rate is initially set to 0.01 and decreased by 0.1 when the training loss does not improve in ten consecutive observations. In Table~\ref{ComparsionOnCTW}, we list the recognition results and resource consumption of different networks. Considering the large number of point convolutions used in Res50, it is reasonable to observe that the reduction in parameters and FLOPs is not very significant. On the other side, we obtain the recognition accuracy improvement by simple replacement.

\begin{table}
\caption{{The comparison of different networks on CTW.}}
\centering \label{ComparsionOnCTW}
\scalebox{0.9}{
\begin{tabular}{|c|c|c|c|c|}
\hline
Model &  FLOPs ($\times 10^9$) &  Storage (MB)      & CER (\%)    \\
\hline
Res50 \cite{yuan2018chinese}   &   \multirow{2}{*}{4.09}       &    \multirow{2}{*}{97.76}       &      21.80  \\
Res50 (Ours)                  &                                  &                             &      20.54           \\
\hline
ParRes50                     &              2.44                    &    58.93    &   19.46  \\
\hline
\end{tabular}}
\end{table}

\subsection{Experiments on MNIST}

\begin{figure*}
\centering
\includegraphics[width=5.3in,height=3.8in]{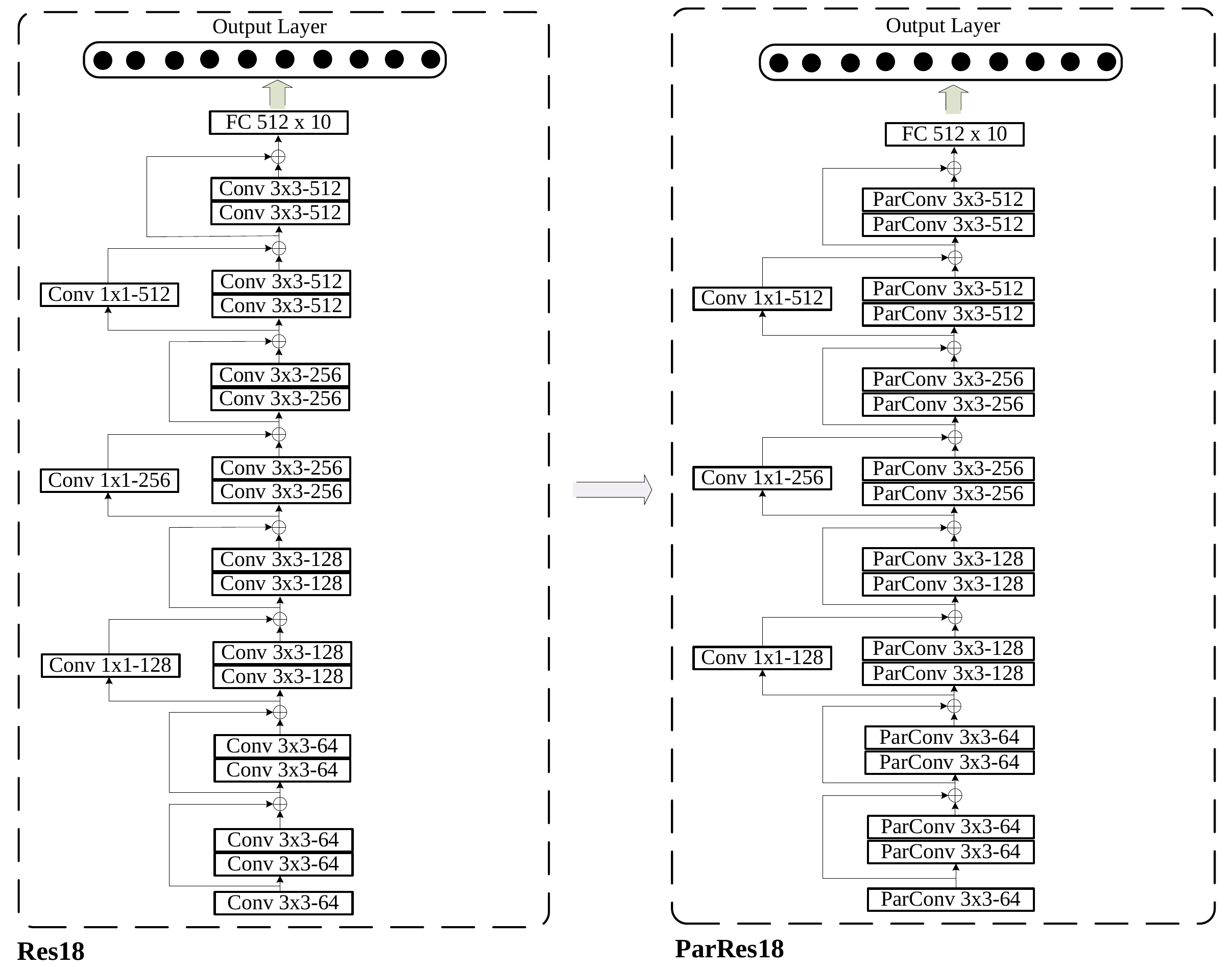}
\caption{Using ParConvs to replace the Convs in Res18.}
\label{Res2ParRes}
\end{figure*}

\begin{table}[h]
\caption{The overall comparison for different networks on MNIST.}
\centering \label{ComparsionOnMNIST}
\scalebox{0.9}{
\begin{tabular}{|c|c|c|c|c|}
\hline
Model &  FLOPs ($\times 10^7$) &  Storage (MB)      & CER (\%)    \\
\hline
AlexNet  \cite{krizhevsky2012imagenet}   &   2.15     &   77.57      &  1.02     \\
\hline
VGG19 \cite{simonyan2014very}               &    27.65     &    148.67    &  0.34 \\
\hline
Res18 \cite{he2016deep}                          &   45.58     &    42.68     &  0.33 \\
\hline
ParRes18                                                   &    4.86   &    4.47      &  0.34 \\
 \hline
\end{tabular}}
\end{table}

In this small dataset, according to the structure of Res18, we build the corresponding compressed network ParRes18 based on the proposed parsimonious convolution with channel multiplier $\omega = 0.5$. Figure.~\ref{Res2ParRes} shows the differences between Res18 and ParRes18. Then, we conduct experiments on one of the most popular datasets: the MNIST dataset that includes 60,000 training images and 10,000 test images. Each image is resized to 28 $\times$ 28 and labeled as a digit (0-9). We first train the three kinds of mainstream neural networks, i.e., AlexNet \cite{krizhevsky2012imagenet}, VGG19 \cite{simonyan2014very}, and Res18. These network prototypes are provided by PyTorch and batch normalization is used for all convolutional layers.  We use the same training criterion to train all networks: the minibatch size is 64, the momentum is 0.9, the weight decay is 0.0001 and the learning rate is set to 0.01. From Table~\ref{ComparsionOnMNIST}, we can observe that compared with Res18, ParRes18 can obtain a $>$9$\times$ acceleration ratio and compression ratio with a similar performance. Besides, it has obvious advantages over AlexNet and VGG19.

\section{Discussion and Conclusion}
\label{sec:con}
\begin{figure*}
\centering
\includegraphics[width=5in,height=2in]{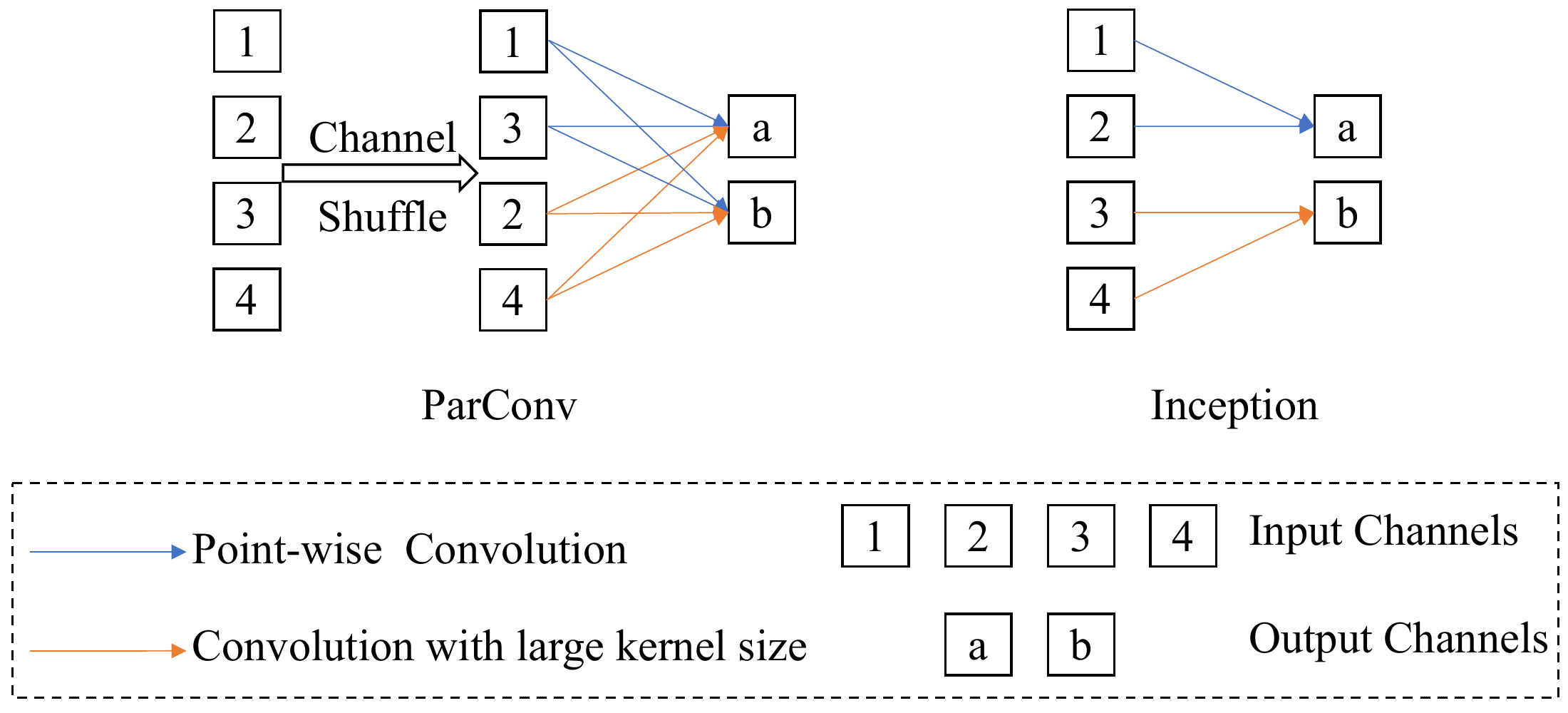}
\caption{A particularly simplified example about ParConv and Inception.}
\label{kernels}
\end{figure*}

As a plug-and-play convolution block, ParConv is proposed to directly replace the standard convolution without other adjustments in the network. Unlike the Inception module in GoogLeNet that uses different kernel sizes in respective paths to extract  multiscale features, the idea of the proposed ParConv derives from the opinion that the convolution filter needs not to have the same spatial correlation on all input channels in an overparameterized CNN.  Therefore, in ParConv, the information can be recovered through the sum (not concatenation)  of corresponding channels extracted from different kernels. In Fig.~\ref{kernels}, a particularly simplified example where the input has four channels and the output has two channels is used to illustrate the different convolutional blocks. We can observe that the ParConv implement the heterogeneous form without additional output channels, which is different from Inception.  Although one path of the proposed ParConv is similar to the inverted residual block in MobileNetv2 \cite{sandler2018mobilenetv2}, they have a different starting point and use. For example, the first pointwise convolution in the inverted residual block of MobileNetv2 increases the number of input channels for extracting more abundant features. However, the ParConv is not perfect, and the accuracy drops when we attempt to achieve the highest compression rate, which is the reason why we still need knowledge distillation. As important work in the future, we will continue to develop more efficient convolution.

In conclusion, we propose a guideline for distilling the architecture and knowledge of pretrained standard CNNs. The proposed algorithm is first verified on offline handwritten Chinese text recognition. In architecture distillation, we invent a parsimonious convolution block (ParConv) to directly replace vanilla convolution without any other adjustments. To further reduce the gap between the baseline CNN and the corresponding compact CNN, knowledge distillation with multiple losses is adopted. Then, by conducting experiments on two additional classification task datasets, CTW and MNIST, we demonstrate that the proposed approach can also be successfully applied on mainstream backbone networks. However, considering that the 1$\times$1 convolutions and depthwise convolutions in PyTorch are relatively slow and that the latest CUDNN library is specially optimized for 3$\times$3 convolutions, the practical speedup is limited. For future work, we will combine other compression and acceleration algorithms to optimize the underlying code and complete the actual deployment.



%



\section*{Acknowledgment}
This work was supported in part by the National Key R\&D Program of China under contract No. 2017YFB1002202, the National Natural Science Foundation of China under Grant Nos. 61671422 and U1613211, the Key Science and Technology Project of Anhui Province under Grant No. 17030901005, and the MOE-Microsoft Key Laboratory of USTC.

\ifCLASSOPTIONcaptionsoff
  \newpage
\fi



\bibliographystyle{IEEEtran}
\bibliography{ref}
\end{document}